\documentclass[]{bytedance_seed}

\usepackage{minitoc}
\usepackage{url}
\usepackage{amssymb}
\usepackage[utf8]{inputenc}
\usepackage{microtype}
\usepackage{booktabs}
\usepackage{pifont} 
\usepackage{multirow}
\usepackage{makecell}
\usepackage{paralist}
\usepackage{xspace}
\usepackage{color}
\usepackage{xcolor}
\usepackage{colortbl}
\usepackage{adjustbox}
\usepackage{hyperref} 
\usepackage[edges]{forest}
\usepackage{tikz} 
\usepackage{caption}
\usepackage{amsfonts}
\usepackage[toc,page,header]{appendix}
\usepackage[utf8]{inputenc}
\usepackage{bbm}
\usepackage{enumitem}
\usepackage{booktabs}
\definecolor{seedc}{RGB}{7, 92, 173}
\usepackage{cleveref}
\usepackage{fancyvrb}
\usepackage{CJKutf8}
\usepackage{threeparttable}
\usepackage[most]{tcolorbox}

\newcommand{\name}[1]{Robix}

\renewcommand{\paragraph}[1]{\vspace{0.1em}\noindent\textbf{#1}}

\usepackage[most]{tcolorbox} 
\usepackage{xcolor}          
\usepackage{lipsum}          
\definecolor{usercolor}{rgb}{0.88, 0.96, 1.0}  
\definecolor{robotcolor}{rgb}{0.95, 0.95, 0.95} 

\newlength{\dialogboxthickness}
\usepackage[most]{tcolorbox}
\usepackage{enumitem}
\usepackage{algorithm}
\usepackage{algorithmic}
\usepackage{multirow}

\usepackage[table]{xcolor} 
\definecolor{lightblue}{HTML}{E6F0FF} 
\usepackage{footmisc}

\title{EmbodiedBrain: Expanding Performance Boundaries of Task Planning for Embodied Intelligence}

\author[]{ZTE NebulaBrain Team}

\abstract{
The realization of Artificial General Intelligence (AGI) necessitates Embodied AI agents capable of robust spatial perception, effective task planning, and adaptive execution in physical environments. However, current large language models (LLMs) and multimodal LLMs (MLLMs) for embodied tasks suffer from key limitations, including a significant gap between model design and agent requirements, an unavoidable trade-off between real-time latency and performance, and the use of unauthentic, offline evaluation metrics. 
To address these challenges, we propose EmbodiedBrain, a novel vision-language foundation model available in both 7B and 32B parameter sizes. 
Our framework features an agent-aligned data structure and employs a powerful training methodology that integrates large-scale Supervised Fine-Tuning (SFT) with Step-Augumented Group Relative Policy Optimization (Step-GRPO), which boosts long-horizon task success by integrating preceding steps as Guided Precursors. 
Furthermore, we incorporate a comprehensive reward system, including a Generative Reward Model (GRM) accelerated at the infrastructure level, to improve training efficiency. For  enable thorough validation, we establish a three-part evaluation system encompassing General, Planning, and End-to-End Simulation Benchmarks, highlighted by the proposal and open-sourcing of a novel, challenging simulation environment. 
Experimental results demonstrate that EmbodiedBrain achieves superior performance across all metrics, establishing a new state-of-the-art for embodied foundation models.
Towards paving the way for the next generation of generalist embodied agents, we open-source all of our data, model weight, and evaluating methods, which are available at https://zterobot.github.io/EmbodiedBrain.github.io.

}

\begin{document}
\maketitle


\section{Introduction}

The pursuit of true Artificial General Intelligence (AGI) hinges on the development of embodied AI, a field that integrates cognitive capabilities with a physical presence. Embodied AI systems are physical agents such as robots that interact with and learn from the complexities of the real world, a necessary step beyond purely computational paradigms. For these agents to be effective, they must master three fundamental capabilities: i) spatial perception to understand their surroundings, ii) task planning to generate effective actions, and iii) a continuous feedback loop of execution and feedback to adapt to the dynamic environment.


Towards building a truly effective AGI, several recent efforts have already extended the capabilities of LLMs/MLLMs to embodied AI, such as Robobrain series, Gemini Robotics.  These approaches typically leverage vision language models (VLMs) as a high-level cognitive module for task planning, decomposing long-horizon tasks into executable subtasks for embodied agents.
However, all of the existing works are exposed to the following limitations:
\begin{itemize}[leftmargin=*]
\item i) \textbf{Huge gap between model and agent}:
As many large-scale LLMs/MLLMs are not intrinsically designed or adapted to the unique constraints and requirements of a physical agent system, they usually get a suboptimal performance, and lack the grounding needed to effectively handle real-world physics, sensor noise, and dynamic changes.
\item ii) \textbf{Difficulty in balancing performance and latency}:
Although many smaller large-scale models are chosen to ensure low latency for real-time operation, these models often exhibit weaknesses in critical capabilities such as instruction following, spatial perception, task decomposing, and so on. This trade-off often compromises the overall reliability and intelligence of the embodied agent. In fact, even the strongest close-source model, such as GPT-O3, also performs poorly when it comes to distinguishing the relation between the target objects.
\item iii) \textbf{Unfair and inauthentic nature of current planning evaluation}:
Existing embodied models are frequently tested exclusively through offline metrics on prebuilt benchmarks, which fail to capture the full complexity and unpredictability of real-world scenarios. This approach neglects the critical aspects of robustness and adaptability, limiting the authenticity and generalization of the evaluation results.
\end{itemize}


We hence propose EmbodiedBrain, the most powerful Embodied vision-language foundation model among both open-sourced and closed-sourced models.
Our EmbodiedBrain addresses the above challenges through a comprehensive framework spanning data, training, and evaluation. 
We began by designing a novel data structure that more closely aligns with the operational needs of embodied agents. Our dataset features a diverse composition specifically curated to reinforce the model’s atomic capabilities in spatial reasoning, instruction following, and task decomposition. 
During model training, we employed a combined methodology of large-scale Supervised Tine-Tuning (SFT) and a large-scale online reinforcement learning approach, Step-Augumented Group Relative Policy Optimization (Step-GRPO), together with a Multimodal Data Rejection Sampling Strategy. Specifically, our proposed Step-GRPO is introduced to enhance the model's capability for long-horizon task planning, which strategically integrates a subset of preceding steps from the long planning sequence to serve as Guided Precursors, boosting the model’s success rate in solving complex planning tasks within the reinforcement learning (RL) framework. 
Besides, we design a comprehensive reward system, incorporating a variety of reward functions and Generative Reward Model (GRM) for which we implemented infrastructure-level accelerations to enhance training efficiency. 
To enable thorough evaluation, we develop a comprehensive evaluation system, including General scenarios, Task Planning, and End-to-End Simulation Benchmarks, covering over 14 benchmarks in total. Besides, a key contribution of our work is the proposal and open-sourcing of a novel end-to-end simulation benchmark, VLM-PlanSim-99, a novel end-to-end simulation benchmark built upon the AI2-THOR environment~\cite{kolve2017ai2}. Featuring a lightweight embodied agent and an effective simulation system, VLM-PlanSim-99 provides a realistic and challenging setting to address the industry’s reliance on inauthentic offline evaluation, enabling robust, real-world-aligned assessment of model performance.


This report provides comprehensive insights for Foundation Robot Brain and is organized as follows:
Section 2 offers a detailed introduction to the architectural design of our proposed EmbodiedBrain. Section 3 presents the specifics of our training data, focusing on the design of the agent-adapted data structure and the composition of data used during the Supervised Fine-Tuning (SFT) and Reinforcement Learning (RL) phases. 
In Section 4, we elaborate on the detailed training strategies employed for both the SFT and RL stages, including the rejection sampling techniques applied in each phase, our novel Step-GRPO reinforcement learning method and the detailed reward system specifically designed for various planning tasks. 
Section 5 is dedicated to our comprehensive evaluation, detailing the selection of general ability benchmarks, spatial perception benchmarks, task planning benchmarks, the design and implementation of our end-to-end simulated Agent Benchmark, ultimately showcasing the superior performance of EmbodiedBrain model. 
Then  Section 6 summarizes the related works in terms of robotic task planning and MLLM reasoning.
Finally, Section 7 summarizes our overall body of work and discusses promising directions for future research.

\section{Architecture}

EmbodiedBrain adopts a modular encoder–decoder architecture designed to unify perception, reasoning, and action planning for complex embodied tasks. Built upon the Qwen2.5-VL framework\cite{bai2025qwen2}, the model comprises three core components: (1) a native-resolution Vision Transformer (ViT) equipped with window-based attention as the vision encoder, (2) a lightweight MLP-based vision–language merger that projects visual features into the embedding space of the large language model, and (3) a decoder-only large language model initialized from Qwen2.5. Although inheriting the strong multimodal foundation of Qwen2.5-VL, EmbodiedBrain is specifically designed for embodied AI scenarios, emphasizing high-precision spatial grounding, temporal reasoning over long video sequences, and causal planning across extended horizons. The architecture seamlessly integrates multi-view images, video frames, and natural language instructions into a unified multimodal token sequence, enabling holistic understanding of dynamic, real-world environments.

The workflow begins with the vision encoder, which employs a Windowed Attention mechanism to efficiently process images at native resolution. Detailed spatial information is captured through 2D Rotary Positional Embedding (ROPE) \cite{su2024roformer}. The resulting visual features are then compressed and aligned with the language domain by an MLP projector. The unified multimodal sequence is subsequently passed to the LLM decoder, which leverages Multimodal RoPE Aligned to Absolute Time to achieve superior temporal understanding. The decoder performs complex reasoning to generate a structured, three-part output comprising a natural language response, a high-level plan, and a sequence of executable actions, allowing direct and nuanced interaction with the environment.

\begin{figure}[htbp]
\centering
\includegraphics[width=5.5in]{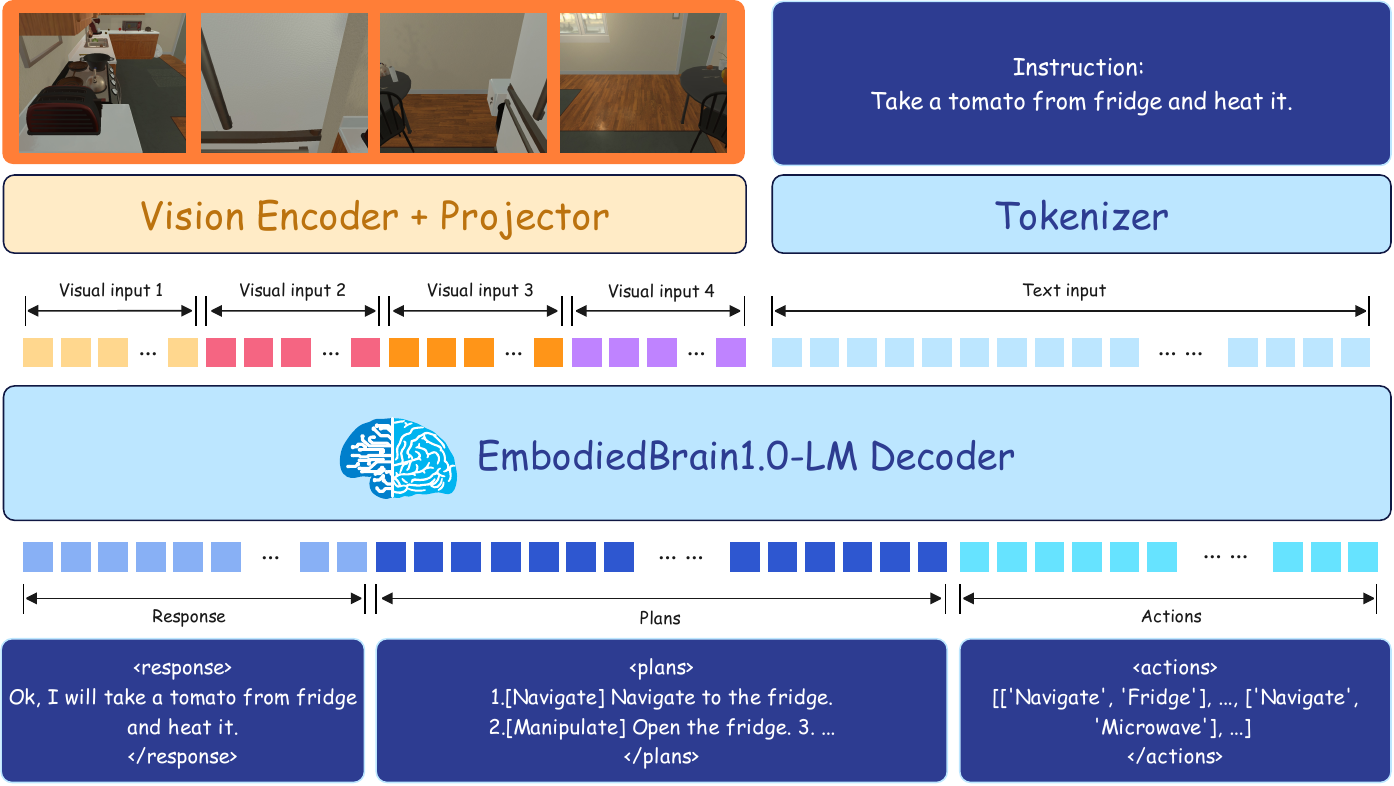}
\caption{\textbf{The Architecture of EmbodiedBrain}: The model processes diverse multimodal inputs, including images of any resolution, long video sequences, and complex language instructions. Visual inputs are handled by a vision encoder and an MLP projector, while textual inputs are tokenized. All inputs are fed into a core LLM decoder, which performs deep reasoning and generates a structured output. The final output consists of three parts: a natural language response (<response>), a step-by-step plan (<plans>), and a sequence of executable actions (<actions>), enabling direct control and interaction within an embodied environment.}
\label{Sec2_framework}
\end{figure}


\section{Training Data}
In this section, we provide an overview of our data framework, detailing the key components that underpin our approach. We begin with a comprehensive introduction to our uniquely designed data structure, which is specifically tailored to support planning-centric agent workflows. This is followed by a description of the diverse composition of data used in both supervised fine-tuning (SFT) and reinforcement learning (RL) stages, including the various sources from which the data is derived—ranging from synthetic generation and human annotation to real-world interaction logs. We also outline the foundational data processing and filtering methodologies employed to ensure data quality, consistency, and relevance, such as noise reduction, action sequence validation, and semantic deduplication. 

\begin{figure}[htbp]
\centering
\includegraphics[width=6 in]{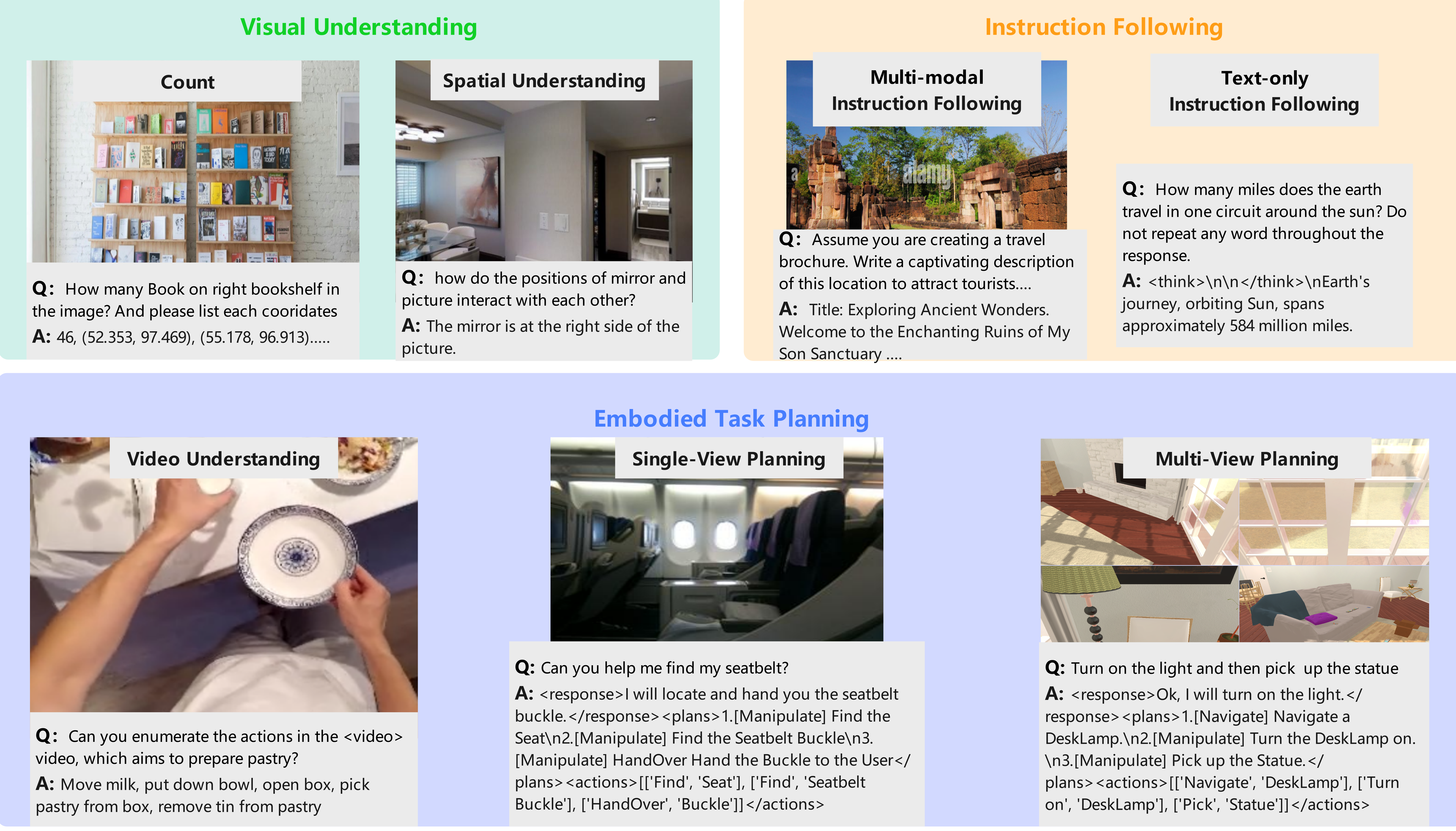}
\caption{ Overview of training data of EmbodiedBrain}
\label{Sec3_datasets}
\end{figure}

\subsection{Data Format Design}

To support embodied intelligence in task-oriented scenarios, we propose a structured data format tailored for planning-centric large language model (LLM) interaction. 
This format is optimized to facilitate agent reasoning, execution, and learning by organizing information into well-defined, machine-interpretable components, which includes a user query, a model-generated response, an explicit structured plan, and the corresponding sequence of low-level actions. By structuring data in this hierarchical and action-oriented manner, we enable more effective alignment between high-level goals and executable steps, making it particularly suitable for embodied agents that require precise planning and grounded action execution. An example is shown below:
    
	\begin{tcolorbox}[colback=gray!10,
			colframe=black,
			width=14cm,
			arc=2mm, auto outer arc,
			title={Answer Format Example of Planning Task},breakable,]	User: \\
            Put the dirty clothes in the washing machine\\
            Assistant:\\
            <response>I will put the dirty clothes in the washing machine</response>\\
            <plans>\\
            1.[manipulate] Locate the dirty clothes in the basket\\
            2.[navigate] Navigate to the basket\\
            3.[manipulate] Pick up the dirty clothes\\
            4.[navigate] Navigate to the washing machine\\
            5.[manipulate] Place the dirty clothes in the washing machine\\
            </plans>\\
            <actions>\\{[[`Search',`Dirty clothes'], [`Navigate',`Basket'], [`Pick',`Dirty clothes'], [`Navigate',`Washing machine'], [`Place',`Dirty clothes',`Washing machine']]}\\
            </actions>	\\
	\end{tcolorbox}\label{data_format}
    
        

The \texttt{<response> </response>} field specifies the textual content of the robot’s spoken utterances while executing a task, enriching the human–robot interaction by providing natural-language feedback, guidance, or status updates that complement visual and tactile modalities. 

The \texttt{<plans> </plans>} field explicitly decomposing the high-level instruction into interpretable planning steps. To ensure modularity and facilitate robotic control, the plan step types are strictly constrained to two categories: \texttt{[Navigate]} and \texttt{[Manipulate]}. Here, \texttt{Navigate} refers to locomotion actions typically executed by the robot's lower limbs, while \texttt{Manipulate} denotes upper-limb operations such as grasping, lifting, or placing objects. This separation enables clear decoupling of motor subsystems and allows the model to invoke corresponding action APIs with greater precision and interpretability.

The \texttt{<actions> </actions>} field encodes the executable steps aligned with the plan, using a compact tuple-based format. Each action is expressed as a binary or ternary tuple (e.g., \texttt{[`Put', `Bread', `Basket']}), which can be directly mapped to agent APIs or robotic control interfaces. Unlike natural language, this structured format ensures consistency, reduces ambiguity, and improves integration with embodied agents.

By bridging high-level reasoning and low-level execution, this format provides a scalable interface for training and evaluating LLMs in embodied planning tasks across both simulated and real-world environments.

\subsection{SFT Data}

\begin{figure}[htbp]
    \centering
    \begin{minipage}[b]{0.38\linewidth}
        \centering
        \includegraphics[width=\textwidth]{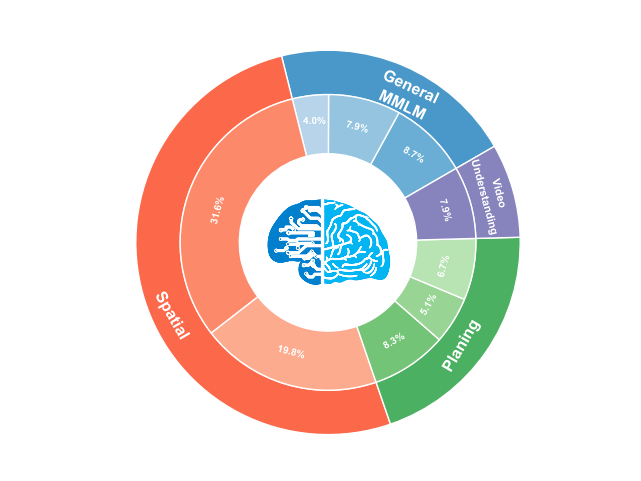}
    \end{minipage}
    \hfill
    \begin{minipage}[b]{0.58\linewidth}
        \centering
        \includegraphics[width=\textwidth]{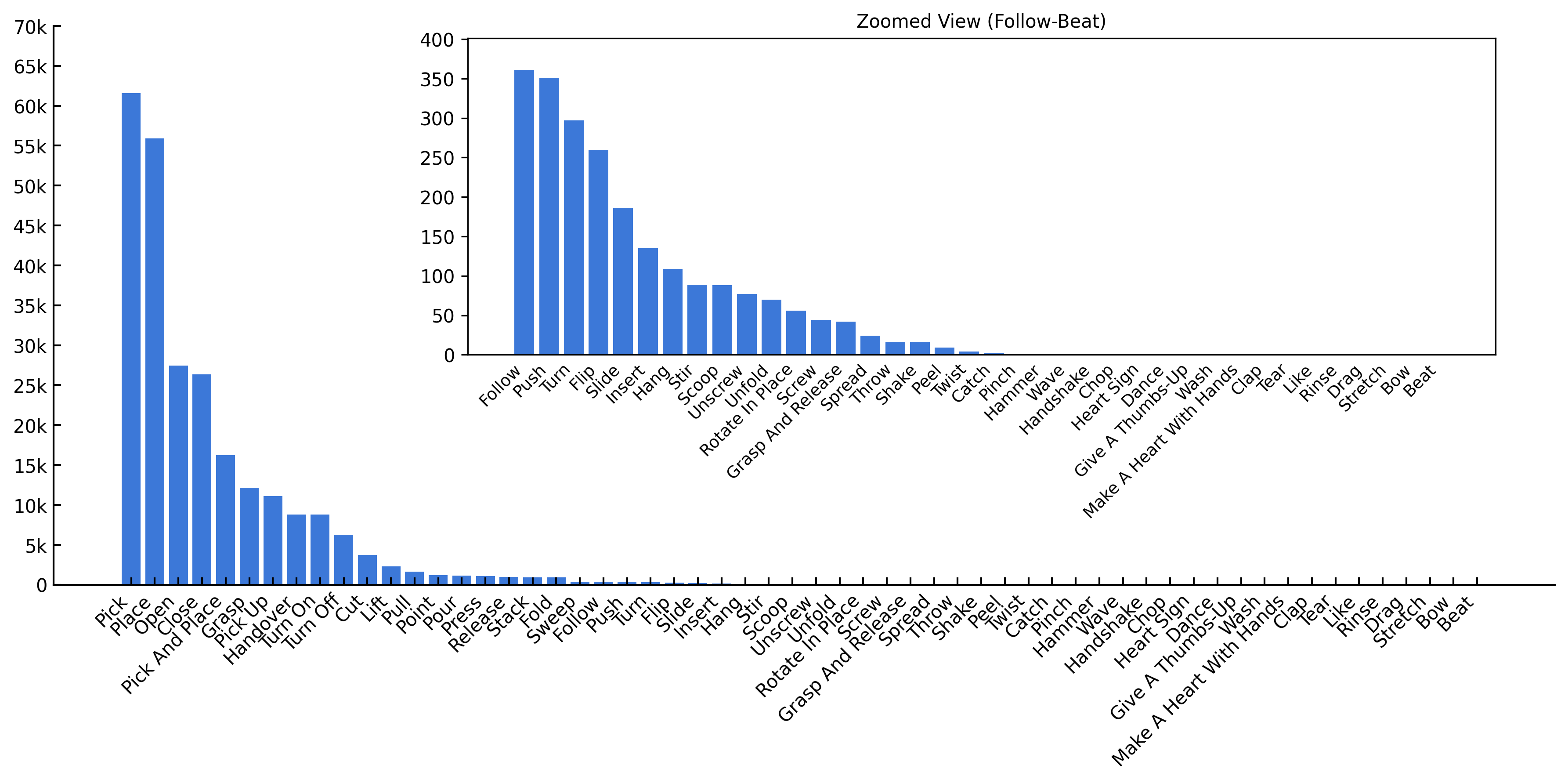}
    \end{minipage}
\caption{ Overall Data Distribution in the SFT Stage and the Distribution of Planning Data for Each Action}
\label{fig:Sec3_datasets_sft_distrubution}
\end{figure}

As shown in Figure~\ref{fig:Sec3_datasets_sft_distrubution}, to equip our EmbodiedBrain with diverse general capabilities, spatial reasoning, and task planning abilities, we constructed datasets that are broad and diverse. The data is divided into four complementary categories. Section \ref{sec:General_LLM_data}
surveys large-scale multi-modal instruction datasets; section \ref{sec:sft_spatial_data} details spatial reasoning and grounding collections; section \ref{sec:sft_planning_data} presents multi-scenario task planning data across simulation and real-world scenarios; and section \ref{sec:sft_video_understanding_data} introduces video understanding data based on real-world scenarios.

\subsubsection{General MLLM data}\label{sec:General_LLM_data} 
\hfill \\

To instantiate the EmbodiedBrain with robust instruction-following, multimodal grounding, and spatial reasoning skills, we meticulously curated a diverse collection of datasets. This comprehensive suite encompasses data for general multi-modal instruction tuning, synthetic constraint-rich prompts, detailed image-grounded dialogues, and complex multimodal perception-generation tasks. In the following subsections, we detail the composition, design motivations, and specific contribution of each dataset to the overall enhancement of embodied intelligence.

(1) \textbf{tulu‑3‑sft‑personas‑instruction‑following}. 
This dataset comprises roughly 30K high-quality samples curated to strengthen the model's capacity for precise instruction adherence and constraint 
satisfaction—capabilities critical for embodied intelligence systems. Covering a broad spectrum of domains, from technical analysis to creative writing 
and factual question answering, it contributes to robust generalization across diverse task settings. Each sample is enriched with an explicit persona 
and 1–3 verifiable constraints, enabling the model to adapt its responses to specific roles while maintaining compliance with task requirements. 
To ensure efficiency and relevance, dialogues exceeding 2,048 tokens were excluded, and rejection sampling was employed to yield a final subset of 10K 
samples, which serves as a core component of our SFT instruction-following training corpus.

(2) \textbf{UltraIF-sft-175k}.
The UltraIF‑sft‑175k dataset comprises 150K single-turn and 25K multi-turn dialogue samples, each augmented with up to three explicit constraints. 
In the multi-turn subset, every turn may either introduce new constraints or carry forward the preceding conversational context. This design is instrumental 
in training the VLM to sustain long-horizon memory in embodied scenarios and to adapt effectively when encountering new or modified instructions mid-task. 
To maintain data quality, dialogues exceeding a predefined length threshold were excluded. For balanced coverage, we uniformly sampled 10K single-turn and 
10K multi-turn dialogues from the full corpus, forming a well-proportioned instruction-following subset within our SFT training pipeline.

(3) \textbf{MM‑IFInstruct‑23k}.
The MM-IFInstruct-23k\cite{ding2025mmifenginemultimodalinstructionfollowing} dataset unifies the strengths of image-grounded dialogue and multimodal perception–generation tasks by pairing textual instructions 
with corresponding visual inputs in onversational form. Each sample is enriched with dual layers of constraints: generation-level requirements 
(e.g., output format, mandated keywords) and perception-level directives (e.g., identifying, describing, or localizing visual elements). This design 
extends beyond the capabilities of language-only instruction-following corpora by introducing explicit perception constraints that compel the model to 
interpret and act upon instructions within visual contexts—such as counting objects, articulating spatial relationships, or describing scene details. 
For embodied intelligence, this multimodal grounding is foundational: it trains the VLM to align linguistic commands with visual observations, enabling 
perception-driven behaviors such as collaborative scene interpretation, spatial reasoning, and object localization. To ensure high data quality, GPT-4o 
was employed to evaluate each sample’s response, with scores ranked from highest to lowest. Rejection sampling was then applied to curate a high-fidelity 
subset of 22K examples for supervised fine-tuning, ensuring that the dataset not only supports accurate instruction adherence but also strengthens the 
model's ability to integrate perception and dialogue in dynamic, real-world environments.

\subsubsection{Spatial data}\label{sec:sft_spatial_data}
\hfill \\
To develop robust spatial reasoning capabilities in our embodied vision–language model (VLM), we curated and processed two complementary datasets: EmbSpatial and pixmo-points. 
Our objective was to ensure high-quality, domain-relevant supervision while maximizing training efficiency.

(1) \textbf{Embspatial}.
The EmbSpatial\cite{du2024embspatialbenchbenchmarkingspatialunderstanding} dataset consists of two subsets: EmbSpatial-Bench (3,640 QA pairs) for evaluation, and EmbSpatial-SFT (25,000 QA pairs) for supervised fine-tuning. We employed a two-stage filtering and categorization pipeline:
\begin{itemize}
    \item \textbf{Baseline Rejection Sampling.} We implemented a rejection sampling procedure by prompting our adopted baseline VLM Qwen2.5VL-7B to generate $k$ responses for each EmbSpatial-SFT question ($k$ is set to eight in this work). These responses were recorded and rigorously compared against the ground truth using a custom natural language analysis tool to identify successful matches.
    \item \textbf{GPT‑4o Verification and Filtering.} Samples for which none of the $k=8$ baseline responses matched the ground truth were subjected to a secondary verification step. GPT-4o was queried with the identical question, and its output was then assessed against the ground truth, effectively serving as a high-fidelity filter for highly ambiguous or complex samples.
\end{itemize}
Based on these evaluations, training samples were classified into four categories:
\begin{itemize}
    \item Category A: All eight baseline responses matched the ground truth.
    \item Category B: A subset of baseline responses matched the ground truth.
    \item Category C: No baseline responses matched, but the answer of GPT‑4o matched.
    \item Category D: Neither baseline responses nor the answer of GPT‑4o matched.
\end{itemize}
Samples from Category D were excluded; a randomly selected subset of Category A was then merged with the complete Category B and C corpora to produce a 50k spatial-reasoning SFT dataset.


(2) \textbf{Pixmo-Points}. For the pixmo-points dataset, we aggregated all annotations associated with the same image into a single multi-turn QA dialogue to improve training efficiency. 
Questions within each dialogue were categorized into:
\begin{itemize}
    \item Counting — e.g., "Please \{collection\_method\} \{label\} in the given image, and answer in natural language. How many \{label\} are in the image?"
    \item Pointing — e.g., "Please write down all the central point coordinates for each \{label\}."
\end{itemize}
Here, \{collection\_method\} and \{label\} were extracted from dataset annotations. We selected \{label\} objects most relevant to embodied intelligence scenarios, sampled the corresponding data, 
and synthesized a 60k-instance training set.

\subsubsection{Planning data}\label{sec:sft_planning_data}
\hfill \\
The Alfred \cite{shridhar2020alfredbenchmarkinterpretinggrounded} dataset is a benchmark for household task understanding and planning, built on the interactive visual environment AI2‑THOR \cite{kolve2022ai2thorinteractive3denvironment}. It defines 25,743 natural language task instructions grounded in five navigation primitives and seven upper‑limb manipulation 
primitives. These instructions span a hierarchy from high‑level goals to fine‑grained sub‑goals. Each task is accompanied by an expert demonstration in the form of a PDDL (Planning 
Domain Definition Language) file, enabling direct execution within the simulation environment to achieve task completion.To construct a task‑planning corpus from Alfred, we designed 
the following pipeline:
\begin{enumerate}
    \item Planning Steps -- Parse PDDL files to generate sub‑task action sequences.
    \item Image \& Detection -- Capture six panoramic images per task in AI2‑THOR and extract object bounding boxes from simulation masks.
    \item Bounding Box Assignment -- Use object IDs from PDDL to distinguish instances and bind boxes to navigable objects in the format \{type: \{id: [x1, y1, x2, y2]\}\}; 
    if no instance is visible, set to [].
    \item Bounding Box Update -- Reset to [] when an object is picked up, as its position changes.
    \item Navigation Update -- For navigation actions, use GotoObject with a bounding box if available; otherwise, use FindObject.
    \item Action Labels -- Tag GotoObject as [Navigate], FindObject as [Map], and all others as [Manipulate].
\end{enumerate}
This process yields a spatially grounded planning dataset for training and evaluating embodied vision–language models.

\subsubsection{Video Understanding data}\label{sec:sft_video_understanding_data}
\hfill \\
The Video Understanding dataset presented in this work is constructed from three foundational sources: Ego4D \cite{grauman2022ego4d}, Epic-Kitchens \cite{damen2022rescaling}, and EgoPlan-IT \cite{chen2023egoplan}. Ego4D and Epic-Kitchens are large-scale egocentric video datasets capturing daily human activities in diverse environments—ranging from domestic and outdoor settings to workplace and kitchen scenarios—providing rich visual and temporal grounding for action observation. Building upon these, EgoPlan-IT leverages large vision-language models to annotate both datasets with high-level task goals and precise temporal boundaries for each atomic action, enabling fine-grained task decomposition and temporal reasoning.

To construct a comprehensive video understanding and planning corpus, we design the following data generation pipeline:
\begin{enumerate}
\item Task and Action Annotation -- Extract high-level task objectives and temporally aligned action segments from EgoPlan-IT annotations, including start and end timestamps for each primitive action.
\item QA Pair Generation -- Generate two types of question-answering (QA) samples:
\begin{itemize}
\item \textit{Retrospective Understanding}: Questions that assess the model’s ability to recognize and describe actions already performed in the video (e.g., "What action was just completed?").
\item \textit{Proactive Planning}: Questions that evaluate the model’s capacity to predict the next action based on current video context and the overall task goal (e.g., "What should be done next to achieve the goal?").
\end{itemize}
\item Multiple-Choice Construction -- For proactive planning QA, use the ground-truth next action as the correct answer, and sample three distractors from temporally nearby actions within the same task trajectory to ensure semantic plausibility.
\item Model-Based Filtering -- Employ Qwen2.5-VL-72B to answer all multiple-choice questions with explicit chain-of-thought reasoning (the detail prompt is provided in \ref{vu_reasoning_prompt}). Only retain question instances where the model generates both a correct answer and coherent justification, ensuring high-quality, interpretable data.
\end{enumerate}

This pipeline yields a dual-purpose video understanding benchmark that not only evaluates perception and recognition capabilities but also advances research in grounded task planning and future action prediction for embodied agents.

\subsection{RL Data}
To optimize policy performance for real-world deployment, the RL phase employs high-fidelity datasets tailored for the large-scale online reinforcement learning stage, which is organized into two complementary streams: i) Spatial Data, refining fine-grained localization and spatial reasoning; ii) Planning Data, fostering the generation of grounded, executable action sequences for long-horizon tasks.

\subsubsection{Spatial data}

The spatial reasoning component of our RL dataset is constructed from two high-quality sources, each selected for its annotation rigor and direct relevance to the spatial challenges faced by embodied agents.

The primary source is a rigorously filtered, non-overlapping subset of 25,000 samples drawn from the EmbSpatial dataset (introduced in Section~\ref{sec:sft_spatial_data}). Recognizing that RL performance is acutely sensitive to reward signal noise, this subset underwent an enhanced, multi-stage verification process detailed in Section~\ref{sec:reject_sampling}. This process employed a consensus-based system integrating the outputs of powerful vision-language models (Qwen2.5VL-72B and Qwen3-32B) with human expert annotation to guarantee exceptional label accuracy. The curated subset is balanced across two fundamental spatial reasoning tasks: \textbf{Target Object Querying} (13,509 samples), which evaluates the model's ability to identify and localize specific objects within a scene, and \textbf{Inter-Object Relationship Querying} (11,491 samples), which assesses its capacity to understand and reason about the spatial relationships between multiple objects. This high-fidelity corpus serves as the cornerstone for refining the model's spatial decision-making under the precise feedback regime of RL.

The secondary source is a specialized subset of the Orsta-Data-47k corpus \cite{ma2025one}, originally curated for VLM post-training via the V-Triune RL system. This dataset, an aggregation from 18 public sources refined through rule-based and difficulty-based filtering, was further processed to retain only the tasks most pertinent to embodied spatial perception. The selected tasks are:
\begin{itemize}
\item \textbf{Object Counting} (1,725 samples): Tasks requiring the model to enumerate objects based on complex, compositional queries (e.g., "How many tiny red objects are there?").
\item \textbf{Object Detection} (8,000 samples): Tasks demanding the identification and localization (via bounding boxes) of all instances of a specified object category within an image.
\item \textbf{Visual Grounding} (4,870 samples): Tasks involving the localization of objects based on rich, descriptive referring expressions (e.g., "Pinpoint a watch with three or more hands").
\end{itemize}

This multi-task spatial dataset provides diverse and challenging perceptual signals, ensuring the model develops a robust and generalizable understanding of its visual environment, crucial for reliable real-world operation.

\subsubsection{Planning data}
To support DAPO-based policy learning, we transformed Self-ReVision \cite{park2025makingvlmsrobotfriendlyselfcritical} and ALFRED \cite{ALFRED20} into visually grounded, atomic-action sequences, thereby enabling precise reward computation for executable planning.

The SelfReVision dataset, originally comprising approximately 26,000 samples, provides GPT-4o-generated high-level task plans in response to natural language instructions within scenes from the Places365 dataset. Recognizing that raw natural language plans are insufficient for policy optimization, we employed Qwen3-32B as a format conversion expert to systematically restructure each plan. Guided by a carefully engineered system prompt, the model was instructed to annotate each planning step with a semantic action tag—such as \texttt{[Navigate]} or \texttt{[Manipulate]} drawn from a predefined ontology. Simultaneously, it generated a corresponding machine-executable action sequence, formatted as a list of verb-object tuples (e.g., \texttt{[["Navigate", "Shelf"], ["Pick", "Book"]]}), ensuring strict alignment with our atomic action vocabulary. The final output for each sample encapsulates the original natural language response, the semantically tagged plan, and the structured action sequence within a unified string. This format, validated for syntactic and semantic consistency, yields a corpus in which the ground-truth policy is unambiguously defined, enabling precise reward assignment during reinforcement learning.

Complementing this, the ALFRED dataset contributes approximately 22,000 samples derived from expert planning domain definition language (PDDL) demonstrations in the AI2-THOR household simulation environment. To adapt this data for real-world policy learning, we transformed each trajectory into a visually grounded, executable plan format compatible with the predefined atomic action space. Each task instance is processed as follows: First, the original low-level PDDL actions (e.g., MoveAhead, RotateRight) are abstracted into high-level semantic steps such as \texttt{[Navigate]} or \texttt{[Manipulate]}, aligning with the modular architecture of the embodied system. Second, six multi-view RGB images are captured per scene at 60-degree intervals, providing 360-degree environmental context for visual grounding. Third, object bounding boxes are extracted from simulation masks and dynamically maintained: when an object is picked up, its bounding box is invalidated to reflect its changed state; when it becomes visible again (e.g., placed on a surface), its box is re-assigned. Fourth, navigation steps are intelligently labeled: If the target object boundary box is visible in the current view, the step is labeled \texttt{[Navigate]}; otherwise, it is labeled \texttt{[Map]}, signaling the agent to perform a search. All manipulation steps (Pick, Place, Open, etc.) are uniformly tagged as \texttt{[Manipulate]}. Finally, each sample is formatted into a structured output containing:
\begin{itemize}
\item A natural language confirmation (\texttt{<response>}),
\item A step-by-step plan with semantic tags (\texttt{<plans>}), 
\item A machine-executable action sequence (\texttt{<actions>}), where each action strictly maps to our predefined atomic action set (e.g., \texttt{["Pick", "Spoon"]}, \texttt{["Place", "Spoon", "Sink"]}).
\end{itemize}

This transformation ensures that each ALFRED sample delivers a visually grounded, structurally executable, and reward-aligned training signal, thereby enabling robust reasoning under partial observability and successful long-horizon task execution in complex, interactive environments.

\section{Training Strategy}

In this section, we present a detailed description of our two-stage training pipeline of EmbodiedBrain, designed to effectively cultivate embodied intelligence in our model.
Starting from a robust vision-language foundation, the first stage focuses on high-quality supervised fine-tuning (SFT), where we employ a multimodal rejection sampling framework to refine the model’s perception and reasoning capabilities.
Building upon this refined initialization, the second stage employs a multi-task reinforcement learning (RL) framework, in which the model is further optimized through environment-based feedback signals across diverse embodied tasks. This stage enhances the generalization of embodied agents, plan over long horizons, and adapt its behavior dynamically in response to real-world constraints. Together, these two stages form a coherent and progressive training paradigm that bridges perception with action, enabling effective learning of complex, goal-driven behaviors in embodied settings.

\subsection{Stage1: Multi-modal Rejection Sampling based SFT}
In this stage, we focus on improving the meta abilities of EmbodiedBrain in a data-efficient manner, including instruction following, spatial understanding and reasoning, and task planning. We mainly perform Rejection Sampling for achieve efficient data utilization, then finetune the VLM with the comprehensively sampled data.
By selectively curating and reweighting training samples based on multimodal consistency and task relevance, this stage ensures that the model acquires robust foundational behaviors aligned with desired action sequences and environmental understanding.

\subsubsection{General Rejection Sampling Strategy}
\label{sec:reject sampoing}



A significant challenge in training powerful embodied AI foundation models is ensuring the quality of the underlying training data. The embspatial dataset, despite its large scale, contains substantial noise and erroneous samples, making manual curation prohibitively expensive. To address this, we developed a systematic two-stage rejection sampling pipeline to programmatically filter a high-fidelity subset for SFT and RL training.

The first stage performs coarse-grained filtering. We leverage the baseline model, Qwen2.5-VL-7B, to generate multiple candidate responses for each sample, which are then scored by a more powerful judge model, Qwen3-30B-A3B-Instruct-2507. If all generated responses are deemed incorrect, the sample is rejected, effectively eliminating obviously flawed or ill-posed queries. Samples that pass the first stage proceed to a second stage of fine-grained verification, designed to validate the accuracy of the original ground-truth labels. Here, we employ the expert model Qwen2.5-VL-72B to generate a high-confidence answer. This generated answer serves as an ‘oracle’ prediction. We then compare it against the original ground truth label. If a significant discrepancy is found, it indicates a high likelihood that the original label is erroneous, and the sample is consequently filtered out. This step is crucial for capturing subtle inaccuracies and label noise that were missed in the first stage.

This hierarchical filtering process effectively removes both overt defects and subtle label errors. The resulting high-quality corpus provides a reliable foundation for training the spatial perception and reasoning capabilities of our EmbodiedBrain model, ensuring that learned behaviors are based on consistent, accurate, and semantically coherent visual-language interactions.

\subsubsection{Cold-start SFT}

\hfill\\
In the cold‑start SFT stage, we explored different mixtures of general MLLM, spatial reasoning, and task‑planning data to identify the ratio that 
maximizes gains in spatial reasoning and planning performance on both our internal and open‑source benchmarks.

\begin{table}[ht]
\centering
\small 
\setlength{\tabcolsep}{6pt} 
\caption{Evaluation of Cold-Start SFT Performance Across Data Mixing Configurations (all values in \%).}
\label{tab:data_ratio_results}
\begin{tabular}{l *{12}{c}} 
\hline
\multirow{2}{*}{\shortstack{Data Amount\\(K)}} & 
\multicolumn{5}{c}{Spatial Reasoning} & 
\multicolumn{6}{c}{Task Planning} \\ 
\cline{2-6} \cline{7-12}
& BLK & CVB & EMb & ERQ & Avg & T/F & SC & EP1 & EP2 & EgT & Avg \\
\hline
30 : 50 : 70 : -     & 83.22 & 78.79 & 71.95 & 41.85 & \textbf{68.95} & 89.75 & 79.98 & 42.89 & 43.73 & 48.92 & \textbf{61.05} \\
30 : 50 : 45.5 : -   & 86.01 & 78.77 & 72.61 & 41.60 & \textbf{69.75} & 90.55 & 82.59 & 42.95 & 42.51 & 49.83 & \textbf{61.69} \\
30 : 50 : 51.5 : -   & 85.31 & 78.18 & 72.39 & 40.85 & \textbf{69.18} & 91.05 & 81.57 & 43.10 & 44.11 & 51.71 & \textbf{62.31} \\
52 : 130 : 51.5 : -  & 89.51 & 79.53 & 74.64 & 39.80 & \textbf{70.87} & 88.35 & 79.73 & 42.14 & 42.21 & 51.90 & \textbf{60.87} \\
52 : 130 : 51.5 : 20 & 87.41 & 80.37 & 73.43 & 39.85 & \textbf{70.27} & 92.10 & 83.23 & 46.95 & 47.91 & 53.00 & \textbf{64.64} \\
\hline
\end{tabular}
\end{table}



As summarized in Table~\ref{tab:data_ratio_results}, we tested five distinct configurations, with amounts (left to right) representing General MLLM, Spatial Reasoning, Task Planning, and Video Understanding data, respectively. While the \texttt{52:130:51.5:-} mix achieved the highest average score in spatial reasoning (70.87\%), it underperformed in task planning (60.87\%). In contrast, the \texttt{52:130:51.5:20} configuration—augmented with 20 units of dedicated planning data—delivered the best overall balance: it maintained strong spatial reasoning performance (70.27\%) while achieving a substantial gain in task planning (64.64\%, the highest among all trials). This configuration notably improved performance on planning subtasks such as execution planning (EP1/EP2) and goal-oriented reasoning (EgT).

Consequently, we adopted the \texttt{52:130:51.5:20} data mixture for the cold-start SFT phase, as it provides the most robust and well-rounded foundation for downstream spatial and planning capabilities.

\subsection{Stage2: Multi-task post-training with Step-GRPO}

\subsubsection{Multimodal Rejection Sampling Strategy}
Rejection sampling in multimodal corpora is more than a filter—it is an active distribution-shaping tool. Its core role is to ensure that every gradient update lands on a sample that is actually “worth learning from.”.

Measurement of the difficulty of multimodal samples remains a fundamental challenge in post-training data curation. For a baseline model whose linguistic prior is already strong, easy instances—those solvable from text alone—are effectively redundant: they deliver minimal gradient signal and merely reinforce existing decision patterns. In contrast, samples of moderate or high complexity compel the model to consult the visual stream, thereby offering genuine opportunities to expand its cross-modal capacity. To identify these valuable cases without relying on external labels, we exploit an intuitive property: challenging multimodal samples should exhibit higher sensitivity to visual information loss—when critical visual content is obscured, the model’s performance degrades more rapidly compared to easier samples. We operationalize this insight through a systematic masking strategy that gradually removes visual information while monitoring prediction stability.

Given an image-text pair $s = (I, Q)$, we systematically probe the model's visual dependence through controlled perturbation experiments. We define a series of masking ratios $\Lambda = \{ \lambda_i \mid \lambda_i = 0.0, 0.1, 0.2, \dots, 0.9 \}$, spanning from the original unmodified image ($\lambda = 0.0$) to heavily degraded versions where $90\%$ of pixels are occluded.

For each masking level $\lambda_i$, we apply the perturbation operation $M(\cdot, \lambda_i)$ that randomly selects and masks the specified proportion of pixels in the original image, yielding $I_{\lambda_i} = M(I, \lambda_i)$. This masking occurs directly in pixel space prior to any feature extraction, thereby simulating realistic scenarios of visual information loss or corruption that might occur in real-world applications.

We then evaluate model performance by feeding each perturbed sample $(I_{\lambda_i}, Q)$ to the multimodal model $\mathcal{M}$ and obtaining the prediction $A_{\lambda_i} = \mathcal{M}(I_{\lambda_i}, Q)$. The correctness of each prediction is assessed using a binary indicator:
\begin{equation}
    \delta_{\lambda_i} = 1[\mathcal{C}(A_{\lambda_i}, A_{\text{gt}})]
\end{equation}
where $\mathcal{C}$ evaluates whether the predicted answer matches the ground truth $A_{\text{gt}}$.

Given the stochastic nature of random masking, we repeat this evaluation process $K=10$ times with independent mask realizations for each masking ratio. The robust accuracy estimate is then computed as:
\begin{equation}
    P_c(\lambda_i) = \frac{1}{K} \sum_{k=1}^{K} \delta_{\lambda_i}^{(k)}
\end{equation}

The critical insight lies in identifying the failure threshold $\lambda_s^*$—the minimal masking ratio where performance drops below an acceptable level. Formally, we define:

\begin{equation}
    \lambda_s^* = \min \{ \lambda_i \in \Lambda \mid P_c(\lambda_i) < \tau \}
\end{equation}
where the threshold $\tau$ (we set $\tau = 0.1$)captures the transition point from reliable to unreliable predictions.

\subsubsection{Step-augumented Group Relative Policy Optimization (Step-GRPO)}

\begin{figure*}[!ht]
    \centering
    \includegraphics[width=0.88\linewidth]{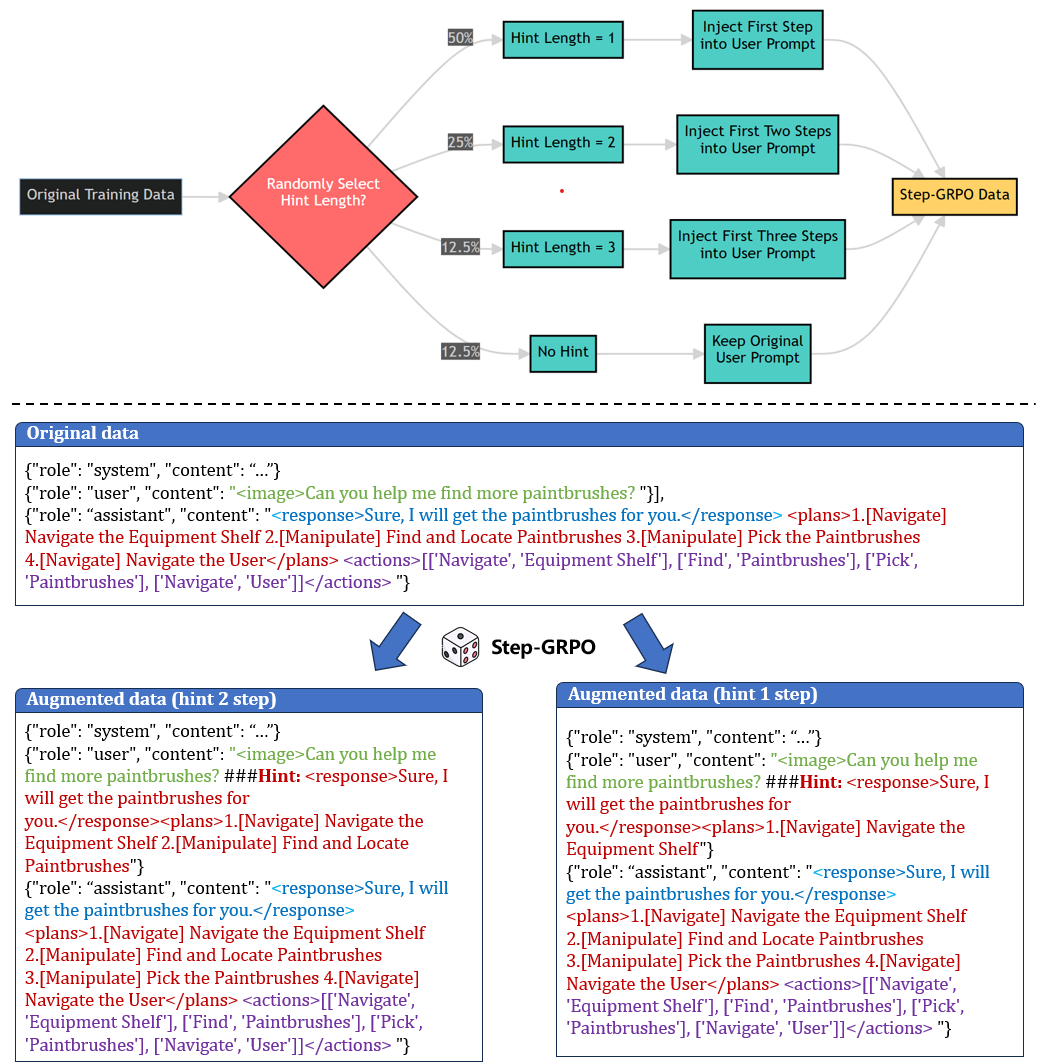}
    \caption{\textbf{Detailed process of the proposed Step-GRPO}}
    \label{fig:step-grpo}
\end{figure*}

The data used in Stage 2 is strategically composed of four complementary categories, each tailored to address distinct aspects of the model's requirements with task-customized reward:

\begin{enumerate}
    \item \textbf{Instruction Following Data}: The instruction-following data used in stage 2 is similar but not overlapping with the SFT stage. The purpose of mixing this part of data in the RL phase is to maintain the model's ability to interpret and act on diverse user inputs, a core component of general capability. The reward logic adheres to a targeted, two-step framework to evaluate whether the model accurately follows instructions. 1) focus exclusively on instruction following tasks while skipping the reward calculation for all other task types. 2) apply the correctness reward referring to the given answer and the ground truth. The score derived from this comparison serving as the reward value for the model’s performance on the instruction following task.

    \item \textbf{Visual Perception Data}: The reward function employs an adaptive, multi-task approach to evaluate visual perception abilities across grounding, detection, and counting tasks. It first dynamically identifies the specific task for each sample by analyzing the structure of the ground truth data, rather than relying on a predefined task type. Once the task is determined, it applies a specialized evaluation metric: for grounding and detection, the reward is calculated based on the spatial accuracy of the predicted bounding boxes, using a weighted Intersection over Union (IoU) to score the overlap with the ground truth boxes. For counting tasks, the reward is simply based on the numerical correctness of the final count, rewarding a precise match with the true value. This task-specific methodology ensures that the model's output is judged by the most relevant criteria for each distinct visual challenge.
    
    \item \textbf{Spatial Perception Data}: The reward logic for spatial perception data relies on an evaluator that uses a two-branch framework, classifying tasks into "multiple-choice questions" or "descriptive questions" by identifying the questioning format. For multiple-choice questions, evaluation strictness is adjusted by answer format (simpler for words/phrases, stricter for sentences) and penalizes excessive verbosity. For descriptive questions (focused on object spatial relationships), it prioritizes semantic understanding—recognizing synonyms, judging logical consistency with antonyms (e.g., swapped prepositions and subject-object positions), and also penalizing unnecessary verbosity. This approach ensures rewards reflect the model’s spatial logic grasp while encouraging concise, semantically accurate outputs.

    \item \textbf{Task Planning Data}: 
    For planning tasks, two types of rewards are designed to optimize two core capabilities of models: output format correctness and planning rationality. To address the former, the model's output is standardized using XML tags—separated into response, plans, and actions (as shown in Example of Section~\ref{data_format}). Firstly, \textbf{a rule-based reward} (ranging from 0 to 1) is implemented via rule-based parsing, which evaluates strict formatting criteria such as tag completeness, proper tag closure, and adherence to a predefined action set. However, relying solely on this rule-based approach risks restricting the model’s action space, often leading to safe but functionally incorrect plans. To mitigate this, \textbf{a Generative Reward Model (GRM)-based reward model} is introduced to assess planning rationality, providing a second score (0 to 1) by judging the alignment between the generated output and the ground truth (here we adopt the Qwen3-30BA3, and the detail prompt is provided in \ref{GRM_prompt}). Crucially, the collaboration of these two rewards presents a key tuning challenge: increasing the weight of the rule-based reward intuitively strengthens format constraints, yet it simultaneously and negatively impacts the stability of the overall training process.

\end{enumerate}
			

The second stage of training focuses on reinforcement learning, more precisely, we proposed a Step-GRPO method, with two overarching objectives that address key limitations of the SFT process in Stage 1 while preserving the model's general capabilities. First, it aims to push the boundaries of the model's performance in complex long-horizon task planning and spatial perception tasks. These tasks typically involve sequential decision-making with extended temporal dependencies and precise interpretation of spatial relationships, which are challenging for SFT to fully optimize given the static nature of labeled data. Second, the RL phase is designed to standardize the model's output format to generate parsable planning steps for downstream applications (e.g., explicitly listing sub-tasks and action sequences). 

The core of Step-GRPO lies in its provision of more informative learning signals, which is inspired by QuestA\cite{li2025questa} and simplifies the problem and mitigates instability in the reinforcement process. As illustrated in Figure~\ref{fig:step-grpo}, under task planning scenarios, we supply hints of random lengths at the planning step level. This approach of problem decomposition stabilizes training and enables more convergent reward dynamics, outperforming both standard GRPO and DAPO in our scenarios.

By the way, to accelerate the training convergence of the Step-GRPO framework integrated with the Grounded Reward Model (GRM), we implemented an asynchronous optimization focused on reward generation. Specifically, the synchronous dependence on Reward Model (RM) inference for scoring introduces significant computational overhead, severely impeding the main Reinforcement Learning (RL) training loop and diminishing overall training throughput. To effectively address this computational constraint, we deploy an asynchronous reward scoring mechanism facilitated by a dedicated multithreaded architecture. This approach successfully decouples the computationally intensive RM inference from the core agent's workload, thereby enabling the parallelization of the reward calculation process. This architectural modification yielded a substantial acceleration of approximately \textbf{20\%} in the end-to-end RL training time, critically, without any observable degradation in the final model's accuracy or performance.

\section{Evaluation}
 
To demonstrate the unprecedented capabilities of EmbodiedBrain and validate its superiority over existing task planning MLLMs, we conducted an extensive and rigorous evaluation across a total of 14 distinct, state-of-the-art multimodal and embodied AI benchmarks. Our multi-faceted assessment is strategically divided into three critical capability domains:
\begin{itemize}[leftmargin=*]
    \item \textbf{General Multimodal Ability (5 benchmarks)}: We assess core understanding and reasoning using established large-scale datasets, including MM-IFEval ~\citep{ding2025mm}, MMMU ~\citep{yue2024mmmu}, MMStar~\cite{chen2024MMStar}, AI2D ~\citep{kembhavi2016AI2D}, and OCRBench~\cite{liu2024ocrbench}.
    \item \textbf{Spatial Perception (4 benchmarks)}: A dedicated evaluation was performed to test the model's spatial reasoning and grounding skills using BLINK \textit{spatial}~\cite{fu2024blink}, cv-bench~\cite{tong2024cambrian},  Embspatial~\cite{du2024embspatial}, and ERQA~\cite{team2025gemini}
    \item \textbf{Comprehensive Task Planning Suite (spanning 3 key categories of 5 benchmarks)}: i) Public Planing benchmarks (Egoplan~\cite{chen2023egoplan}, Egoplan2~\cite{qiu2024egoplan2}, and EgoThink~\cite{cheng2024egothink}); ii) A specialized Internal Planning benchmark designed for fair, challenging long-range sequence evaluation (Internal Planing); iii) Our newly proposed Internal End-to-end Simulation Benchmark (VLM-PlanSim-99).
\end{itemize}
This thorough scope ensures a complete picture of EmbodiedBrain's performance across generalized reasoning, perception, and embodied action planning.
Next we will first introduce each benchmark across above three critical capability domains, then give a comprehensive conclusion on the model performance of EmbodiedBrain.

\subsection{Evaluation of General Ability}

General understanding and common sense reasoning are critical capabilities for foundational models, especially in embodied and multimodal contexts. To assess the general reasoning ability of EmbodiedBrain, we conducted a comprehensive evaluation using the OpenCompass suite~\citep{duan2024vlmevalkit}, which integrates five representative benchmarks: \textbf{MM-IFEval}, \textbf{MMStar}, \textbf{MMMU}, \textbf{AI2D}, and \textbf{OCRBench}.

\begin{itemize}
    \item \textbf{MM-IFEval} focuses on multimodal instruction following, testing a model’s ability to understand and execute complex commands grounded in visual input. This benchmark is central to our evaluation, as improving instruction-following performance is a key objective for EmbodiedBrain. We aim to enhance its responsiveness to natural language directives in real-world scenarios, such as cooking, cleaning, or object manipulation.

    \item \textbf{MMStar} evaluates general multimodal reasoning across diverse domains, including perception, spatial understanding, and temporal inference. It serves as a broad test of the model’s ability to integrate visual and textual information coherently.

    \item \textbf{MMMU} (Multimodal Massive Multitask Understanding) probes knowledge-intensive reasoning across multiple disciplines, such as science, history, and mathematics, using both textual and visual cues. It helps ensure that our model retains strong factual and analytical capabilities.

    \item \textbf{AI2D} targets diagram understanding, requiring models to interpret structured visual information such as charts, graphs, and scientific illustrations. This benchmark is crucial for assessing the model’s ability to process schematic representations and spatial relationships.

    \item \textbf{OCRBench} measures optical character recognition and text-based reasoning in images, such as reading labels, signs, or embedded text in natural scenes. It ensures that the model can extract and reason over textual content in visual environments.
\end{itemize}

While \textbf{MM-IFEval} is our primary focus—given its relevance to instruction-following in embodied tasks, we also aim to maintain competitive performance across the other benchmarks. Our goal is to improve EmbodiedBrain’s instruction-following ability without sacrificing its general reasoning and perception capabilities compared to baseline models.




\subsection{Evaluation of Spatial Perception}
We rigorously assess the spatial reasoning capabilities of the EmbodiedBrain model using four commonly adopted and complementary benchmarks in the embodied AI domain:

\begin{itemize}
    \item \textbf{BLINK}: Following the protocol established by preceding works (e.g., RoboBrain 2.0), our evaluation focuses exclusively on the two core spatial-relation tasks: spatial understanding and depth perception. This specific selection isolates the model's ability to interpret and infer basic geometric and positional relations solely from linguistic cues, making it a foundation test of spatial grounding.

    \item \textbf{CV-Bench}: This benchmark assesses detailed spatial reasoning by presenting complex queries that necessitate the understanding of 3D object properties, positions, and inter-object relationships within a static scene. CV-Bench critically tests the model’s competence in interpreting visual scenes with high geometric and relational precision, ensuring it can process fine-grained spatial data.

    \item \textbf{EmbSpatial}: EmbSpatial is specifically designed to measure egocentric spatial relationships within dynamic 3D embodied environments. By requiring the model to reason about spatial arrangements from a first-person perspective, it directly evaluates skills essential for real-world robotic applications, such as dynamic object manipulation and successful navigation planning in unfamiliar spaces.

    \item \textbf{ERQA}: ERQA represents the most challenging test of end-to-end multimodal reasoning for embodied agents. It demands that the model integrate high-level spatial, visual, and linguistic cues to answer complex questions grounded in simulated real-world scenarios, thereby providing a comprehensive, integrated measure of an agent's general embodied intelligence.
\end{itemize}



\subsection{Evaluation of Planning}

\subsubsection{Public Planing benchmark}


We evaluate the planning capabilities of EmbodiedBrain using both public planning benchmarks and internal evaluation protocols. These benchmarks are designed to assess how well a model can understand task goals, reason about future actions, and generate coherent multi-step plans in embodied environments.

\begin{itemize}
    \item \textbf{EgoPlan Benchmark Series}~\cite{chen2023egoplan,qiu2024egoplan2} are embodied reasoning benchmarks that focus on goal-oriented planning in egocentric 3D environments. They test a model’s ability to interpret visual scenes, infer task objectives, and produce sequential action plans that align with human-like behavior. The benchmark includes diverse tasks such as cooking, cleaning, and object manipulation, making it a strong indicator of practical planning ability.

    \item \textbf{EgoThink}~\cite{cheng2024egothink} is a comprehensive benchmark that evaluates both planning and reasoning capabilities in egocentric settings. It includes a wide range of tasks that require temporal reasoning, spatial understanding, and goal decomposition. EgoThink emphasizes the integration of perception and cognition, challenging models to not only plan actions but also justify them based on environmental cues.

\end{itemize}



\subsubsection{Internal Planing benchmark}

Current public embodied intelligence planning benchmarks predominantly focus on \textbf{short-term, single-step action planning}, where models predict the next immediate action based on the current state. They largely lack a comprehensive evaluation for the critical ability of \textbf{long-range, complex action sequence planning}. 

To systematically validate our model's robust capability for planning across multiple steps and complex tasks, we have developed the \textbf{Internal Planning Benchmark}. This internal suite is specifically designed to  measure the model's capacity for generating long-sequence plans.

The benchmark incorporates diverse testing scenarios sourced from both real-world data from the \textbf{Sharerobot} platform and simulated environments within \textbf{AI2THOR}. For each test case, the powerful \textbf{Qwen30B-A3B} model was utilized to generate the high-quality \textbf{Ground Truth (GT) planning steps}. All competing models are prompted under a uniform system setting to perform inference on the simulation benchmarks. 


\begin{tcolorbox}[title=\textbf{Pseudocode for Internal Planning Benchmark Evaluation}, colback=gray!5, colframe=black!80, sharp corners, boxrule=1pt]

\begin{algorithmic}[1]
    \STATE \textbf{Input:} $\mathcal{P} = \{\text{Pred}_1, \dots, \text{Pred}_m\}$ (Predicted Action List)
    \STATE \textbf{Input:} $\mathcal{G} = \{\text{GT}_1, \dots, \text{GT}_n\}$ (Ground Truth Action List)
    \STATE \textbf{Evaluator:} $\text{GPT-5-Mini}_{\text{eval}}$ (Action Matching Model)
    \STATE \textbf{Algorithm:} $\text{Hungarian}(\cdot)$ (Max Bipartite Matching)
    \STATE \textbf{Algorithm:} $\text{LCS}(\cdot)$ (Longest Common Subsequence)

    \COMMENT{\textbf{Preprocessing: Filter out Low-Level Actions}}
    \STATE $\mathcal{P} \leftarrow \text{FilterLowLevel}(\mathcal{P})$ 
    \STATE $\mathcal{G} \leftarrow \text{FilterLowLevel}(\mathcal{G})$
    \STATE $m \leftarrow |\mathcal{P}|$, $n \leftarrow |\mathcal{G}|$

    \COMMENT{\textbf{Step 1: Construct Cross-Matching Matrix $M$}}
    \STATE $M \leftarrow \text{Matrix}(\text{rows}=m, \text{cols}=n, \text{init}=0)$
    \FOR {$i = 1$ to $m$}
        \STATE $\text{pred\_act} \leftarrow \mathcal{P}[i]$
        \STATE $\text{match\_info} \leftarrow \text{GPT-5-Mini}_{\text{eval}}(\text{pred\_act}, \mathcal{G})$
        \FOR {each $j$ in $\text{match\_info}$}
            \STATE $M[i, j] \leftarrow 1$  \COMMENT{Mark match}
        \ENDFOR
    \ENDFOR

    \COMMENT{\textbf{Step 2: Calculate Match Quantity Metrics (M-QUANTITY)}}
    \STATE $\text{QuantityScore} \leftarrow \text{Hungarian}(M)$ 
    \STATE $\text{P}_{\text{Quant}} \leftarrow \text{QuantityScore} / m$
    \STATE $\text{R}_{\text{Quant}} \leftarrow \text{QuantityScore} / n$
    \STATE $\text{F1}_{\text{Quant}} \leftarrow \frac{2 \cdot \text{P}_{\text{Quant}} \cdot \text{R}_{\text{Quant}}}{\text{P}_{\text{Quant}} + \text{R}_{\text{Quant}}}$

    \COMMENT{\textbf{Step 3: Calculate Match Order Metrics (M-ORDER)}}
    \STATE $\text{OrderScore} \leftarrow \text{LCS}(\mathcal{P}, \mathcal{G}, M)$ \COMMENT{LCS constrained by $M$}
    \STATE $\text{P}_{\text{Order}} \leftarrow \text{OrderScore} / m$
    \STATE $\text{R}_{\text{Order}} \leftarrow \text{OrderScore} / n$
    \STATE $\text{F1}_{\text{Order}} \leftarrow \frac{2 \cdot \text{P}_{\text{Order}} \cdot \text{R}_{\text{Order}}}{\text{P}_{\text{Order}} + \text{R}_{\text{Order}}}$

    \COMMENT{\textbf{Step 4: Aggregate Results}}
    \STATE \textbf{Return:} $\{\text{P}_{\text{Quant}}, \text{R}_{\text{Quant}}, \text{F1}_{\text{Quant}}\}, \{\text{P}_{\text{Order}}, \text{R}_{\text{Order}}, \text{F1}_{\text{Order}}\}$
\end{algorithmic}
\end{tcolorbox}

\textbf{\seedblue{The Internal Planning Benchmark Construction Piplie}}

\begin{enumerate}
    \item \textbf{Preprocessing} \\
    We extract the action list from the \texttt{<actions>} field of the inference results, specifically excluding low-level locomotor actions (e.g., \texttt{find}, \texttt{navigate}). The filtered predicted action list (Pred Action List) and the GT Action List are then submitted to a GPT-5-Mini evaluation model for scoring, ensuring objective assessment.

    \item \textbf{Construction of the Cross-Matching Matrix} \\
    To precisely quantify the correspondence between the predicted and GT sequences, we first construct a Cross-Matching Matrix $M(m \times n)$, where $m$ and $n$ are the lengths of the predicted and GT action pairs, respectively, initialized to all zeros. We iterate through the predicted action pairs, inputting one pred action pair along with the entire GT action pair list into the GPT-5-Mini evaluator. The evaluator returns the matching information (potentially one-to-many matches), and the corresponding position in matrix $M$ is set to 1 if a match is confirmed.

    \item \textbf{Metric Calculation} \\
    The final evaluation metrics are composed of Match Quantity Metrics and Match Order Metrics, both presented as Precision, Recall, and F1-Score:
    \begin{itemize}
        \item \textbf{Action Pair Match Quantity Score:} The maximum number of matches (the match score) is computed from the cross-matching matrix $M$ using the Hungarian Algorithm via the SciPy library.
        \item \textbf{Action Pair Match Order Score:} The order score is calculated using the Longest Common Subsequence (LCS) algorithm, where the length of the LCS serves as the match score.
    \end{itemize}

    The metrics are calculated as follows, with \texttt{Score} representing the derived match score (either quantity or order):

    \begin{align*}
        \text{Precision} &= \frac{\displaystyle \text{Score}}
                                {\displaystyle \text{Predicted Actions}} \\[6pt]
        \text{Recall}    &= \frac{\displaystyle \text{Score}}
                                {\displaystyle \text{Ground Truth Actions}} \\[6pt]
        \text{F1-Score}  &= \frac{\displaystyle 2 \cdot \text{Precision} \cdot \text{Recall}}
                                {\displaystyle \text{Precision} + \text{Recall}}
    \end{align*}
\end{enumerate}

\subsubsection{Internal End-to-end Simulation Benchmark: VLM-PlanSim-99}
To rigorously assess the task planning capabilities of our proposed model, EmbodiedBrain, we developed a comprehensive, physically-grounded evaluation framework. Standard language-based metrics, such as BLEU or ROUGE, are insufficient for task planning as they only measure textual similarity and fail to capture the executability, efficiency, and logical coherence of a generated plan in the real world. Therefore, our methodology transcends static textual comparisons by validating EmbodiedBrain's output through end-to-end execution in a high-fidelity 3D simulation environment. This approach allows for a holistic evaluation of the model's capacity to integrate visual perception, commonsense reasoning, and sequential decision-making.

Our evaluation framework is comprised of two core components: (1) a novel, meticulously curated benchmark dataset, which we term \textbf{VLM-PlanSim-99}, containing guaranteed-executable ground-truth plans; and (2) a three-stage automated pipeline that processes EmbodiedBrain's raw output and validates its performance within the AI2-THOR simulator.

\textbf{\seedblue{The VLM-PlanSim-99 Benchmark Construction}}

The foundation of our evaluation is the \textbf{VLM-PlanSim-99} benchmark, a high-quality test dataset designed for grounded task planning. In contrast to procedurally generated datasets, VLM-PlanSim-99 is a manually curated collection of 99 distinct household task instances. For this benchmark, human experts meticulously selected each sample, ensuring a diverse and representative range of common domestic scenarios.

Each instance in the dataset is comprised of two key components: (1) a high-level task goal described in natural language (e.g., "Rinse a mug and place it in the coffee machine"), and (2) the corresponding initial environmental images that provide the necessary visual context. Subsequently, our team manually annotated a detailed, step-by-step list of sub-tasks for each goal, creating a gold-standard ground-truth plan. A critical aspect of our curation process was the rigorous validation of every annotated plan. Each sequence of sub-tasks was executed directly within the AI2-THOR simulator to guarantee its executability and correctness. This verification ensures that every ground-truth plan is not only logically coherent but also physically achievable, providing a reliable basis for our evaluation.

\textbf{\seedblue{Embodied-AI Task Planning Verifying Framework}}

\begin{figure}[th] 
    \centering  
    \subfloat[Overall look of three-stage pipeline for VLM-based task planning assessment.]
    {   \begin{minipage}[t]{0.45\textwidth}
            \centering          
            \includegraphics[width=\textwidth]{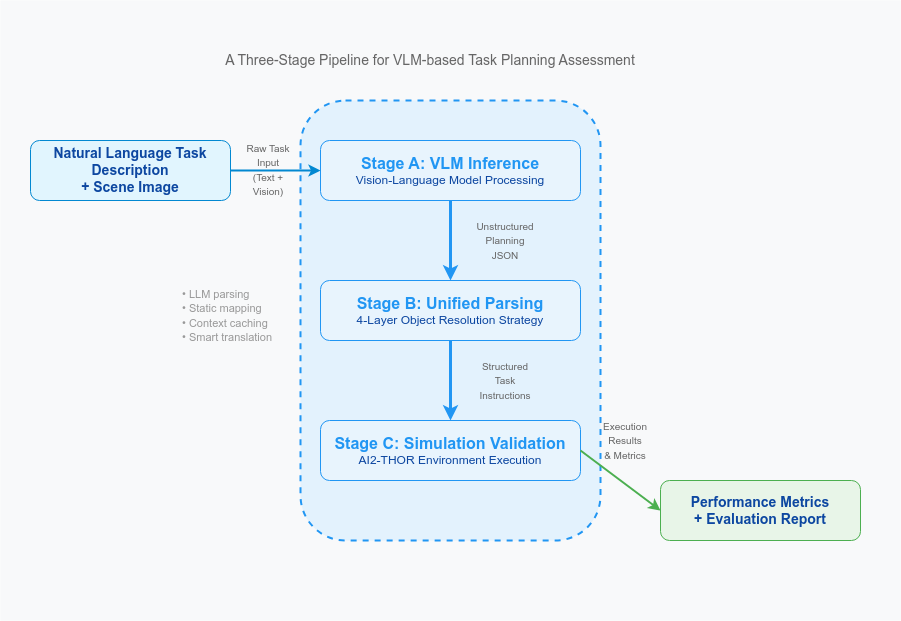} 
        \end{minipage}}
    \subfloat[Detailed architecture of the Stage C Task Verification Framework]
    {   \begin{minipage}[t]{0.5\textwidth}
            \centering     
            \includegraphics[width=\textwidth]{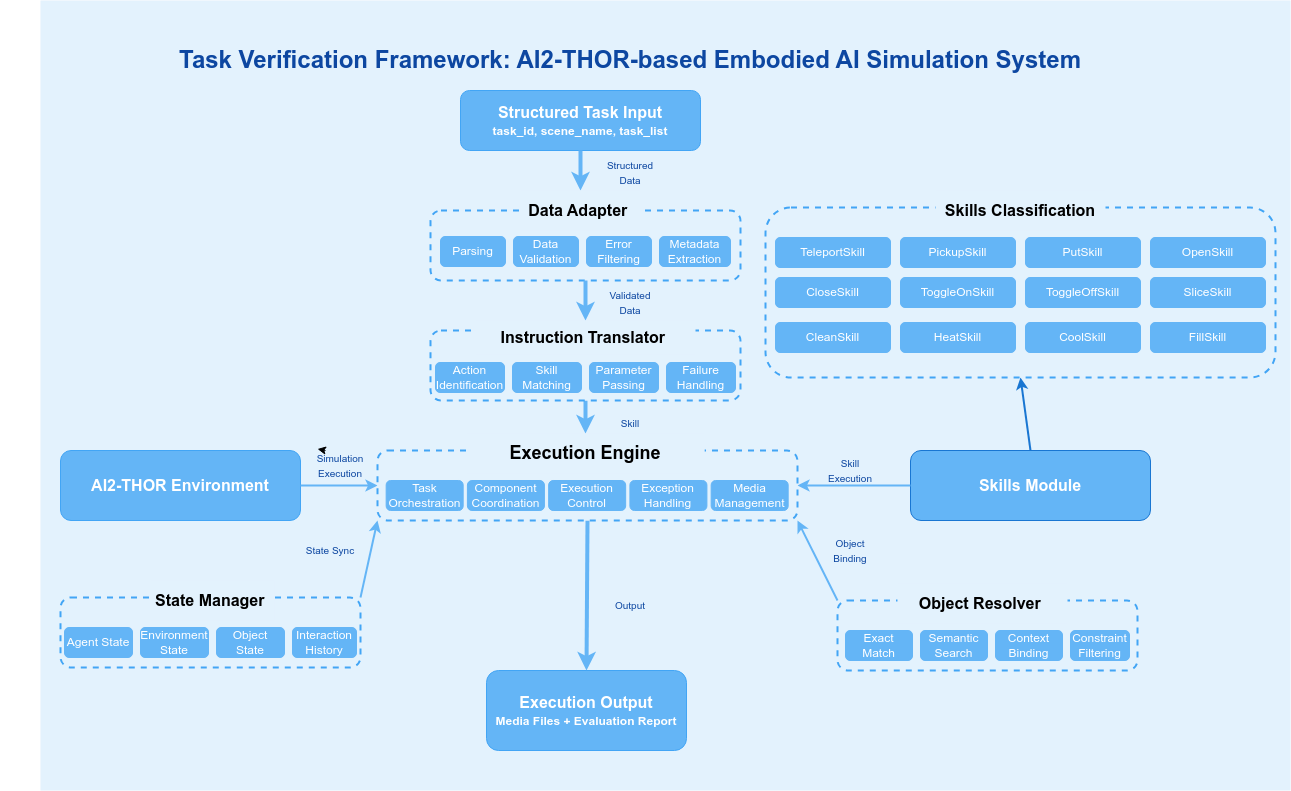} 
        \end{minipage}}
    \caption{The three-stage pipeline for VLM-based task planning assessment (left) takes a natural language description and a scene image as input, processes it through Inference, Parsing, and Simulation stages, and outputs quantitative performance metrics. The Stage C Task Verification Framework (right) contains structured data input through adaptation, translation, and execution within the AI2-THOR environment, managed by a central execution engine.} 
    \label{fig:vlm_assessment_pipeline} 
\end{figure}

Our automated pipeline connects EmbodiedBrain's abstract planning with concrete execution in the AI2-THOR simulator to quantitatively measure its performance, primarily through the Task Success Rate. As depicted in Figure~\ref{fig:vlm_assessment_pipeline}, the framework operates in three sequential stages.

First, in Stage A (VLM Inference), EmbodiedBrain processes the raw multi-modal input—a natural language task description and a scene image—to generate an initial, unstructured plan. This raw plan is then fed into Stage B (Unified Parsing), where it is translated into machine-executable steps. This crucial stage employs our 4-Layer Object Resolution Strategy (encompassing LLM parsing, static mapping, context caching, and smart translation) to produce standardized, structured instructions. Finally, in Stage C (Simulation Validation), these structured instructions are executed in the AI2-THOR environment. The results from the simulation are aggregated to generate a final evaluation report, providing key performance metrics, including the overall task success rate.

\textbf{\seedblue{Stage C: Simulation Validation Framework}}

As depicted in Figure~\ref{fig:vlm_assessment_pipeline}, Stage C is underpinned by a comprehensive verification framework designed for robust, closed-loop task execution in the AI2-THOR environment. This system is engineered to translate abstract structured instructions from Stage B into concrete, executable actions.

The process initiates when the \texttt{Structured Task Input} enters the \texttt{Data Adapter}, which acts as a vital pre-processing layer, performing \texttt{Data Validation} and \texttt{Error Filtering}. The validated data then proceeds to the \texttt{Instruction Translator}, which functions as the semantic interpreter. It performs \texttt{Action Identification} and \texttt{Skill Matching} to map the instruction to a specific high-level skill from the \texttt{Skills Classification} library (e.g., \texttt{PickupSkill}, \texttt{HeatSkill}).

The entire workflow is governed by the central \texttt{Execution Engine}, which is responsible for high-level \texttt{Task Orchestration} and runtime \texttt{Exception Handling}. When executing a skill, the engine signals the \texttt{Skills Module}. This module, in turn, utilizes the sophisticated \texttt{Object Resolver} to accurately bind the action to a specific in-game object, employing a cascade of methods including \texttt{Semantic Search} and \texttt{Context Binding}. The \texttt{Execution Engine} sends direct execution commands to the \texttt{AI2-THOR Environment} and maintains synchronization with the \texttt{State Manager}, which meticulously tracks \texttt{Agent State} and \texttt{Object History}. This architecture enables robust component coordination and media management, ultimately producing the final \texttt{Execution Output}, which includes qualitative media files and the quantitative evaluation report.

\subsection{Conclusion of Experimental Results}

\begin{table}[t]
\resizebox{\textwidth}{!}{
\begin{tabular}{lcc>{\columncolor{lightblue}}c cc>{\columncolor{lightblue}}c}
\toprule
\textbf{Benchmark} & \textbf{\begin{tabular}[c]{@{}c@{}} Qwen 2.5\\ VL 7B \end{tabular}} & \textbf{\begin{tabular}[c]{@{}c@{}} RoboBrain\\ 2.0 7B \end{tabular}} & \textbf{\begin{tabular}[c]{@{}c@{}}EmbodiedBrain\\ 7B\end{tabular}} & \textbf{\begin{tabular}[c]{@{}c@{}} Qwen 2.5\\ VL 32B \end{tabular}} & \textbf{\begin{tabular}[c]{@{}c@{}} RoboBrain\\ 2.0 32B\end{tabular}} & \textbf{\begin{tabular}[c]{@{}c@{}}EmbodiedBrain\\ 32B\end{tabular}} \\
\midrule
\multicolumn{7}{c}{\textbf{Multimodal General Ability}} \\ \hline
MM-IFEval    & 39.56 & 30.82 & \textbf{43.61} & 46.66 & 39.75 & \textbf{46.98} \\
MMSTAR       & \textbf{62.27} & 59.40 & 62.17 & 64.70 & \textbf{65.8} & 65.40 \\
MMMU         & 51.33 & 48.67 & \textbf{52.67} & 60.00 & \textbf{60.89} & 60.44 \\
AI2D         & 82.55 & 81.83 & \textbf{82.61} & \textbf{85.37} & 85.23 & 84.39 \\
OCRBench     & \textbf{785} & 757 & 783 & 740 & 732 & \textbf{741} \\
\bottomrule
\hline
\multicolumn{7}{c}{\textbf{Spatial Perception}} \\ \hline
BLINK        & 58.74 & 62.94 & \textbf{88.11} & 73.43 & 68.53 & \textbf{87.41} \\
CV-Bench     & 62.03 & 62.97 & \textbf{80.69} & 75.57 & 68.27 & \textbf{83.64} \\
EmbSpatial   & 51.76 & 52.12 & \textbf{75.04} & 67.39 & 62.95 & \textbf{77.03} \\
ERQA         & 41.00 & 42.50 & 41.75 & 44.61 & \textbf{45.11} & 43.50 \\
\textbf{Average} & 53.38 & 55.13 & \textbf{71.40} & 65.25 & 61.22 & \textbf{72.90} \\
\bottomrule
\hline
\multicolumn{7}{c}{\textbf{Embodied Task Planning}} \\ \hline
EgoPlan-Bench  & 41.30 & 36.73 & \textbf{49.10} & 51.11 & 46.83 & \textbf{54.66} \\
EgoPlan-Bench2 & 38.63 & 33.54 & \textbf{49.58} & 49.81 & 49.96 & \textbf{57.11} \\
EgoThink       & 52.13 & 44.92 & \textbf{53.54} & \textbf{56.75} & 49.33 & 53.92 \\
Internal Planing &30.00 & 68.30 & \textbf{85.80} & 28.30 & 75.90 & \textbf{90.50} \\
VLM-PlanSim-99 & 23.2 & 21.21 & \textbf{31.31} & 25.25 & 24.24 & \textbf{46.46} \\
\hline
\end{tabular}
}
\caption{Performance of EmbodiedBrain on 14 distinct benchmarks compared to prior models. The highest score in each benchmark is highlighted in bold within each group.}
\label{tab:benchmarks}
\end{table}


\textbf{The EmbodiedBrain architecture achieves state-of-the-art performance across comprehensive embodied scenarios.} Our evaluation framework was rigorously designed to assess the EmbodiedBrain models against leading multimodal large language models (MLLMs), with a specific focus on three critical capabilities: general multimodal reasoning, 3D spatial perception, and embodied task planning. As summarized in Table~\ref{tab:benchmarks}, the results consistently demonstrate that EmbodiedBrain establishes a new performance standard, particularly in the specialized domain of embodied task planning. Next, we analyze the experimental results of three domains in detail.

 \textbf{\seedblue{Outstanding General Multimodal Capabilities}}

As detailed in Tables~\ref{tab:benchmarks}, the proposed EmbodiedBrain models exhibit robust and consistent performance when benchmarked against the Qwen2.5 VL baseline model across five foundational benchmarks: MM-IFEval, MMStar, MMMU, AI2D, and OCRBench. This consistency is crucial, as it confirms that our specialized training regimen successfully preserved the general-purpose reasoning and extensive knowledge base of the baseline model, preventing catastrophic forgetting.

It is pertinent to note a minor discrepancy: the 32B variants of MLLMs generally underperform relative to their 7B counterparts on the OCRBench benchmark. This anomaly is hypothesized to stem from the model’s reliance on more complex, hybrid reasoning pathways at the larger scale, which can introduce ambiguity or repetition during the focused OCR evaluation process.

Most importantly, EmbodiedBrain-7B demonstrates superior proficiency in instruction-following tasks, achieving a score of \textbf{43.61\%} on the challenging MMIF-Eval benchmark, which is actually a huge challenge for models with smaller sizes, e.g., RoboBrain2.0-7B gets only 30.82\% in MMIF-Eval. This represents a substantial \textbf{10.24\% relative improvement} over the Qwen2.5VL baseline, unequivocally highlighting its enhanced ability to parse and execute complex, multimodal instructions. These collective results validate the effectiveness of our training strategy, confirming that embodied-specific enhancements can be achieved without compromising foundational generalization ability.

\textbf{\seedblue{Superior Spatial Perception Capabilities}}

The assessed spatial benchmarks necessitate a high degree of 3D spatial perception and reasoning. This capability is not an intrinsic strength of standard 2D-centric multimodal large language models (MLLMs). Consequently, direct application of models like Qwen2.5-VL-7B to these tasks resulted in notably poor performance, confirming their inherent limitations in understanding and reasoning within 3D spatial contexts.

In stark contrast, the EmbodiedBrain models consistently and significantly outperforms the Qwen2.5-VL baseline across all of the spatial perception tasks. This performance gap validates that our specialized training strategy is highly effective in developing the crucial 3D spatial reasoning capabilities required for real-world embodied navigation and interaction.

Furthermore, when benchmarked against other contemporary embodied MLLMs of comparable size, such as RoboBrain 2.0, EmbodiedBrain achieves substantial relative improvements, establishing its leadership in this domain: i) The EmbodiedBrain-7B model demonstrates relative performance gains of \textbf{39.99\% (BLINK), 28.14\% (CV-Bench), and 43.98\% (EmbSpatial)} compared to RoboBrain2.0-7B. ii) Similarly, the EmbodiedBrain-32B variant achieves relative improvements of \textbf{27.55\% (BLINK), 22.51\% (CV-Bench), and 22.37\% (EmbSpatial)} over RoboBrain2.0-32B.
This consistent and significant advantage across multiple benchmarks indicates the strong generalization and superior robustness of EmbodiedBrain when faced with diverse visual contexts and complex spatial challenges.

\textbf{\seedblue{Advanced Embodied Task Planning Performance}}

The most significant performance gains are demonstrated in the domain of embodied task planning, which is the primary focus of the EmbodiedBrain architecture. We evaluated proficiency across multiple metrics, including egocentric planning, long-horizon internal planning, and physically-grounded simulation.
\begin{itemize}[leftmargin=*]
    \item \textbf{Egocentric and Abstract Planning}: 
    Evaluation on the EgoPlan Series and EgoThink benchmarks confirms that EmbodiedBrain delivers state-of-the-art results within its parameter size category, securing substantial gains over comparable open-source models. The EgoPlan Series emphasizes multi-step planning from first-person (egocentric) perspectives, whereas EgoThink focuses on abstract reasoning and intention modeling within complex embodied scenarios. Specifically, on EgoPlan-Bench2, EmbodiedBrain demonstrates significant relative performance gains over RoboBrain 2.0: 47.82\% for the 7B variant and 14.31\% for the 32B variant. These substantial uplifts indicate a high proficiency in parsing spatial-temporal cues and translating them into logical, executable plans. Furthermore, in EgoThink, the EmbodiedBrain models achieve 19.19\% (7B) and 9.30\% (32B) relative improvements compared to RoboBrain 2.0, convincingly establishing their strong abstract reasoning and robust logical generation abilities.
    
    \item \textbf{Long-Horizon Planning}: 
    The models demonstrate a significant leap in internal planning capabilities, which is crucial for addressing the critical challenge of context accumulation in complex, long-horizon tasks. Compared to the Qwen2.5-VL baseline, EmbodiedBrain-7B and EmbodiedBrain-32B achieve substantial absolute F1 score improvements of 0.558 and 0.622 over their respective Qwen2.5-VL counterparts. This wide performance margin highlights the efficacy of EmbodiedBrain's training advancements in managing complex task dependencies over extended durations. Furthermore, when benchmarked against RoboBrain 2.0, EmbodiedBrain-7B and EmbodiedBrain-32B still exhibit superior performance, delivering absolute F1 score improvements of 0.175 and 0.146 over the RoboBrain2.0 7B and 32B variants, respectively. This compelling performance across multiple baselines solidifies the superior long-horizon planning ability of the EmbodiedBrain architecture, with the 32B variant achieving the highest F1 score of 0.905 in the benchmark.
    
    \item \textbf{Physically-Grounded Task}: To definitively evaluate the efficacy of our architecture in real-time, physically-grounded execution, we utilized the custom VLM-PlanSim-99 benchmark, which specifically measures the Task Success Rate in a simulation framework. As shown in the final rows of Table~\ref{tab:benchmarks}, our models achieve state-of-the-art performance by a commanding margin: Our EmbodiedBrain-7B model achieves a 31.31\% task success rate, clearly outperforming the \texttt{Qwen 2.5 VL 7B} (23.2\%) and \texttt{RoboBrain 2.0 7B} (21.21\%). The performance advantage is even more dramatic at the 32B scale. Our EmbodiedBrain-32B model scores 46.46\%, effectively doubling the performance of the next-best baselines, \texttt{Qwen 2.5 VL 32B} (25.25\%) and \texttt{RoboBrain 2.0 32B} (24.24\%). This result unequivocally demonstrates that our training methodology is highly effective, successfully adapting large pre-trained VLMs to the complex, specific domain of embodied task planning and significantly surpassing other leading models in the critical measure of execution success.

\end{itemize}

\section{Related Work}

\subsection{Robotic Task Planning}
To achieve the capability of solving complex, long-horizon tasks in open environments, Multimodal Large Language Models (MLLMs) have recently been introduced in various works, advancing robotic task planning by grounding high-level instructions in perceptual context.
Initial approaches, such as COME-robot, VILA, and REPLAN, leverage powerful models like GPT-4V to generate executable plans directly from raw visual observations. They then iteratively refine these plans based on environmental feedback, thereby improving overall robustness through situated reasoning and failure recovery. While these works successfully construct a simple system for task planning, their efficacy fundamentally relies on the intrinsic, basic planning capabilities of the underlying MLLM.
Subsequent efforts have focused on enhancing this foundation. For instance, the Robobrain series introduced incremental pre-training and post-training techniques, such as Continued Pre-Training (CPT) and Supervised Fine-Tuning (SFT), to specifically augment the model's core task planning abilities. This augmentation, coupled with more customized agent systems, leads to a significant elevation of system-level task planning competence.
More recently, works including Embodied-Reasoner, Gemini Robotics-ER, Robobrain2, and Robix have further advanced the state-of-the-art. These methods incorporate Reinforcement Learning (RL) post-training to specifically enhance the model's Chain-of-Thought (CoT) capabilities, achieving notable improvements in the agent’s complex, system-level planning performance.

\subsection{MLLM Reasoning}

Benefiting from the rapid advancement in reasoning capabilities within both the MLLM and LLM domains, Embodied Foundation Models have seen repaid improvements in task planning performance, largely driven by enhanced Chain-of-Thought (CoT) abilities. This subsection, therefore, reviews the current CoT efforts relevant to MLLMs.
Early work concentrated on augmenting models' CoT capabilities through Supervised Fine-Tuning (SFT) based on manually curated, rule-based, or ultra-large model-generated CoT corpora, e.g., LLava-COT, Mulberry, etc.
Following the inspiration of Deepseek-R1, subsequent research introduced the Group Relative Policy Optimization (GRPO) algorithm to reinforce CoT generation capabilities using data from domains like mathematics and code. This approach is exemplified by works such as VLM-R1, Visual-RFT, Vision-R1, etc.
More recent studies have expanded the application of RL beyond mathematics and code, exploring its usage on diverse data to enhance model accuracy and generalization for specific tasks, as demonstrated by approaches like Perception-R1 and Curr-Reft. 
Concurrently, a surge of refinement and improvement efforts targeting the core GRPO algorithm is also emerging, often focusing on stability, sample efficiency, and applicability to long-sequence generation tasks, such as VAPO, DAPO, and Dr.GRPO, etc.

\section{Conclusion and Future Works}
In this work, we introduced EmbodiedBrain, a powerful vision-language foundation model specifically engineered to address key limitations in current Embodied AI. We established a comprehensive framework defined by a novel agent-aligned data structure and an advanced training pipeline including Step-GRPO, which successfully enhances the model's ability to solve complex, long-horizon tasks by utilizing preceding planning steps as guided precursors. We validated the superior performance of EmbodiedBrain through a rigorous, three-part evaluation system—covering General, Task Planning, and End-to-End Simulation benchmarks and contributed a novel, open-sourced simulation benchmark to the community, establishing a new reliable evaluation benchmark.

Looking forward, our future work will focus on scaling EmbodiedBrain to handle multi-agent cooperative tasks and exploring domain randomization techniques to ensure even more seamless deployment across a wider variety of real-world robotic platforms.

\clearpage
\section{Core Author List}

\textbf{\seedblue{Contributors}}

\begin{flushleft}
\begin{tabular}{@{}lllll@{}}
Ding Zou & Feifan Wang & Mengyu Ge   & Siyuan Fan  & Zongbing Zhang \\
Wei Chen & Lingfeng Wang & Zhongyou Hu & Wenrui Yan  & Zhengwei Gao   \\
Hao Wang & Weizhao Jin &  Yu Zhang     &  Hainan Zhao &  Mingliang Zhang                 \\
Xianxian Xi & Yaru Zhang &  Wenyuan Li   
\end{tabular}
\end{flushleft}

\vspace{10pt}

\textbf{\seedblue{Project Leader}}


\begin{flushleft}
Zhengguang Gao\footnote[2]{Corresponding Authors. Email: gao.zhengguang@zte.com.cn, zhu.yurui@zte.com.cn, zou.ding@zte.com.cn, and 
wang.feifan2@zte.com.cn} \quad Yurui Zhu\footnotemark[2]
\end{flushleft}

\clearpage

\section{Appendix}

\subsection{Evaluation Examples}
\subsubsection{Spatial Reasoning}

In embodied intelligence, spatial reasoning is a critical capability that allows agents to perceive and interpret the arrangement of objects in their environment, 
understand relational dependencies, and identify available free space for action. Vision-Language Models (VLMs) provide a unified framework for these tasks by 
combining multimodal perception with structured reasoning, enabling agents to ground linguistic queries in visual-spatial contexts. This integration supports 
essential functions such as object localization, relational inference, and affordance-based planning.  

\begin{figure*}[ht]
    \centering
    \includegraphics[width=0.88\linewidth]{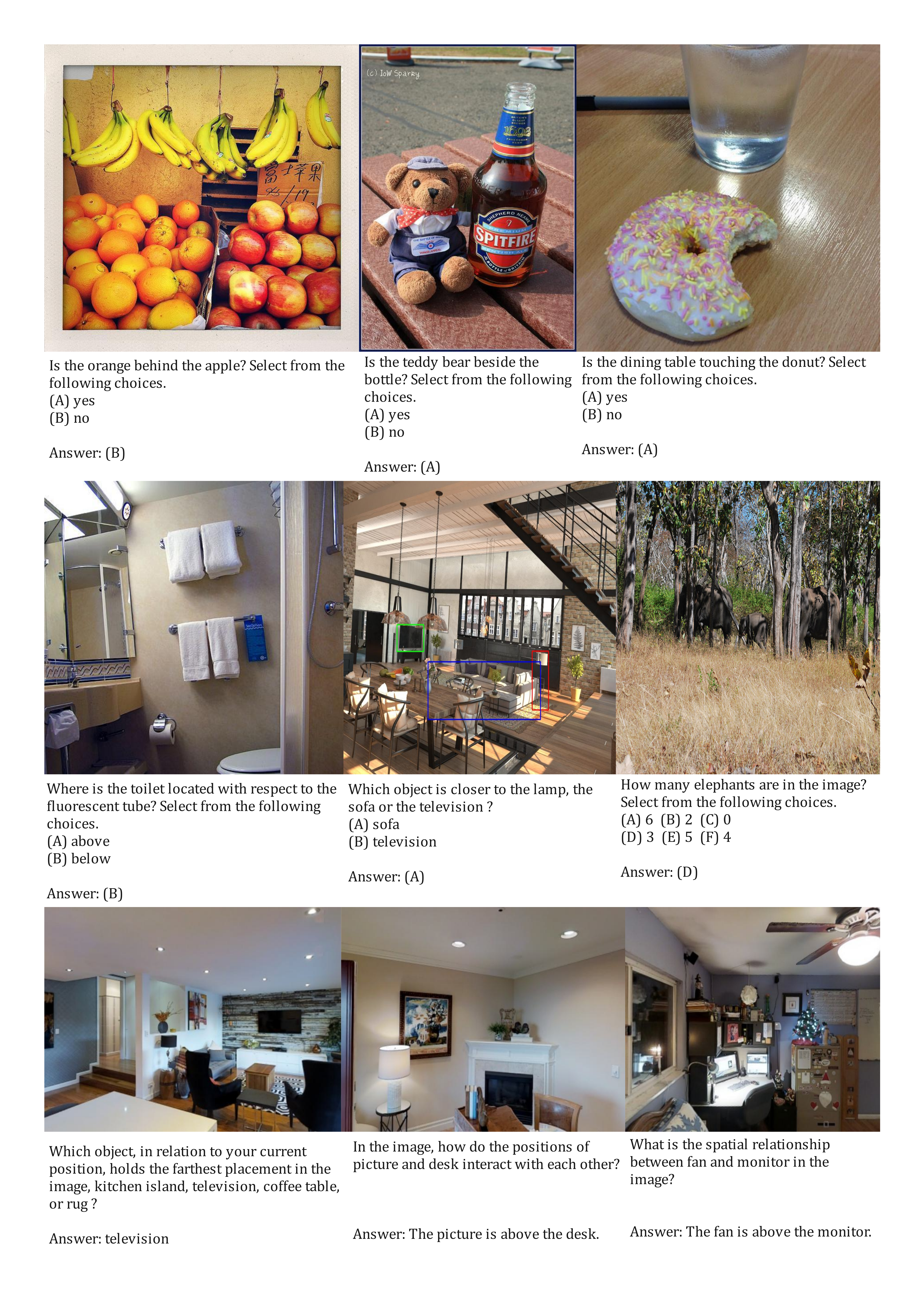}
    \caption{\textbf{Spatial Examples of EmbodiedBrain.}}
    \label{fig:spatial_benchmark_samples_01}
    \vspace{-1.5em}
\end{figure*}

To evaluate these capabilities, we conducted a series of controlled tests designed to probe our model's ability to localize objects, reason about their relative 
positions, and identify spatial relationships across diverse scenarios. The results, summarized in Figure \ref{fig:spatial_benchmark_samples_01}, demonstrate that the model consistently 
produced accurate and contextually appropriate answers across a range of spatial reasoning queries. These included binary judgments of relative placement, 
assessments of proximity, and recognition of hierarchical spatial relations (e.g., “above,” “below,” or “beside”). The model also exhibited robustness in 
handling more complex prompts that required integrating multiple spatial cues, such as determining the farthest object from a given reference point or 
identifying the interaction between two objects in a scene.  

Overall, the outcomes presented in Figure \ref{fig:spatial_benchmark_samples_01} highlight the model's strong performance in spatial reasoning tasks. The ability to correctly resolve both 
simple and compound spatial queries indicates that the VLM effectively integrates visual grounding with linguistic abstraction. This suggests that such models 
are well-suited for embodied intelligence applications where accurate spatial understanding is essential, including navigation, manipulation, and human-robot 
interaction. Future work will focus on extending these evaluations to more dynamic and unstructured environments, thereby further validating the scalability 
and generalizability of VLM-based spatial reasoning.  

\subsubsection{Planning}

\textbf{\seedblue{Public and Internal Planing benchmarks}}

\begin{figure*}[ht]
    \centering
    \begin{subfigure}{0.45\linewidth}
        \centering
        \includegraphics[width=\linewidth]{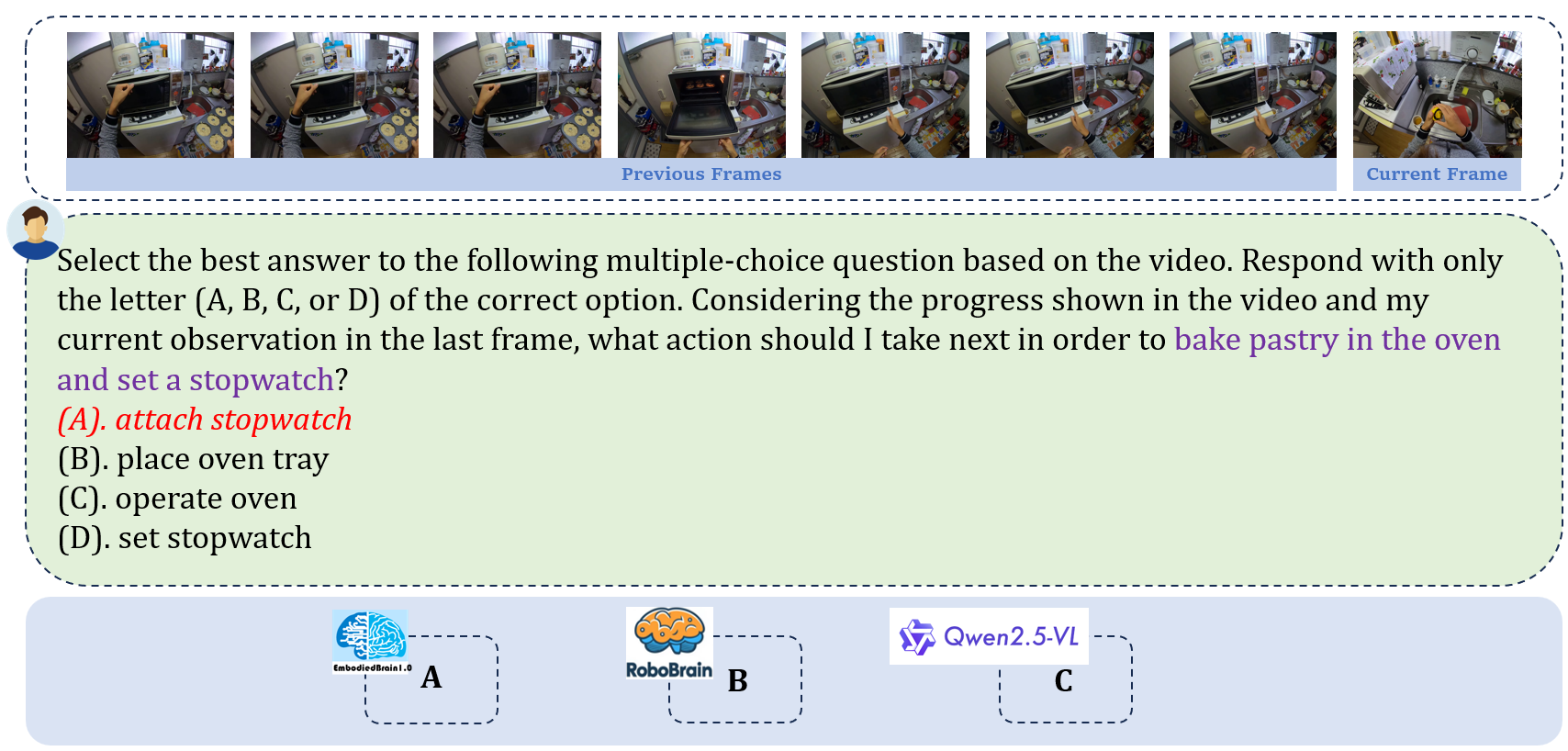}
        \caption{Egoplan Example 1}
        \label{fig:ego1}
    \end{subfigure}
    \hfill
    \begin{subfigure}{0.45\linewidth}
        \centering
        \includegraphics[width=\linewidth]{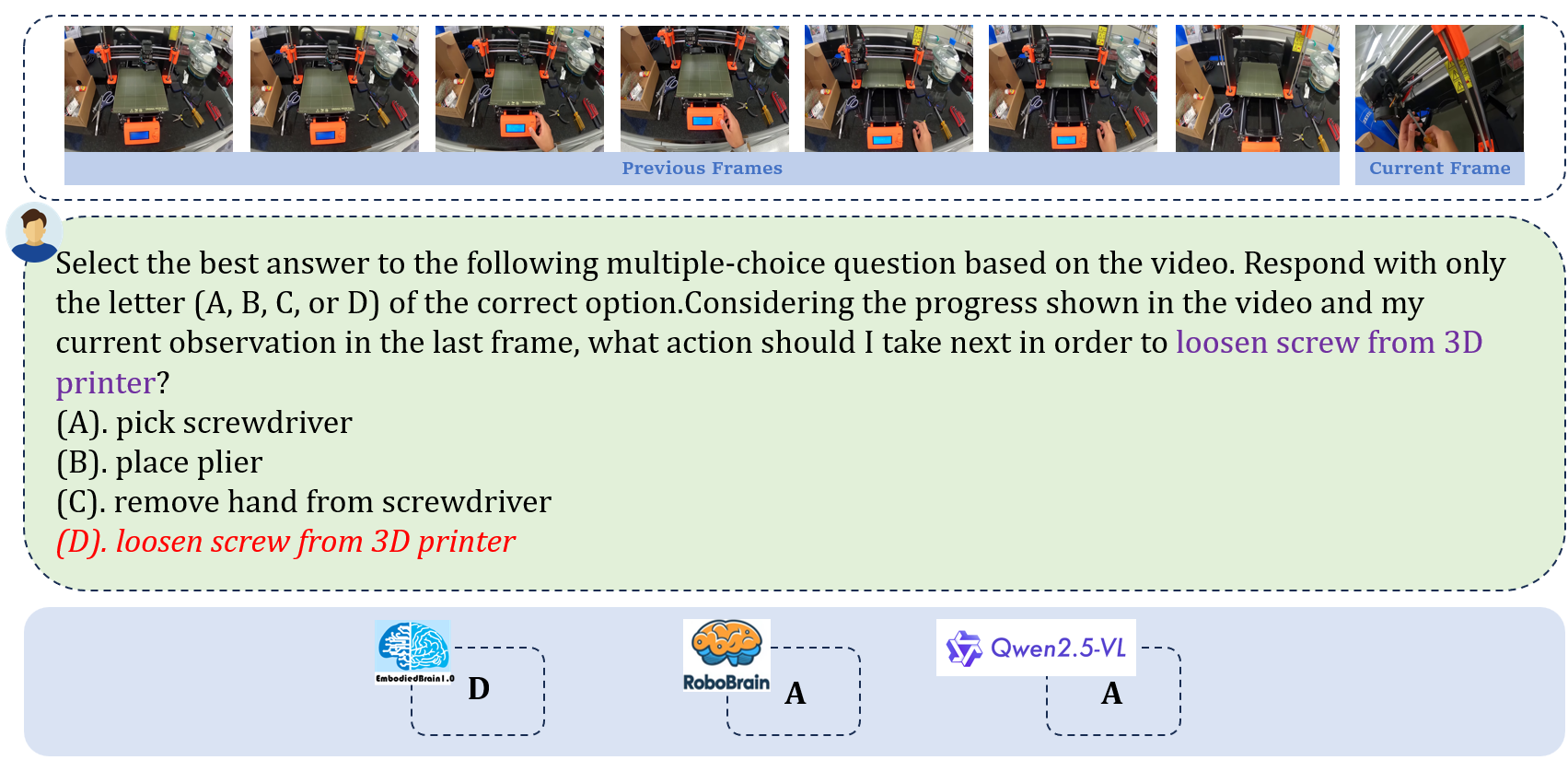}
        \caption{Egoplan Example 2}
        \label{fig:ego2}
    \end{subfigure}

    \begin{subfigure}{0.45\linewidth}
        \centering
        \includegraphics[width=\linewidth]{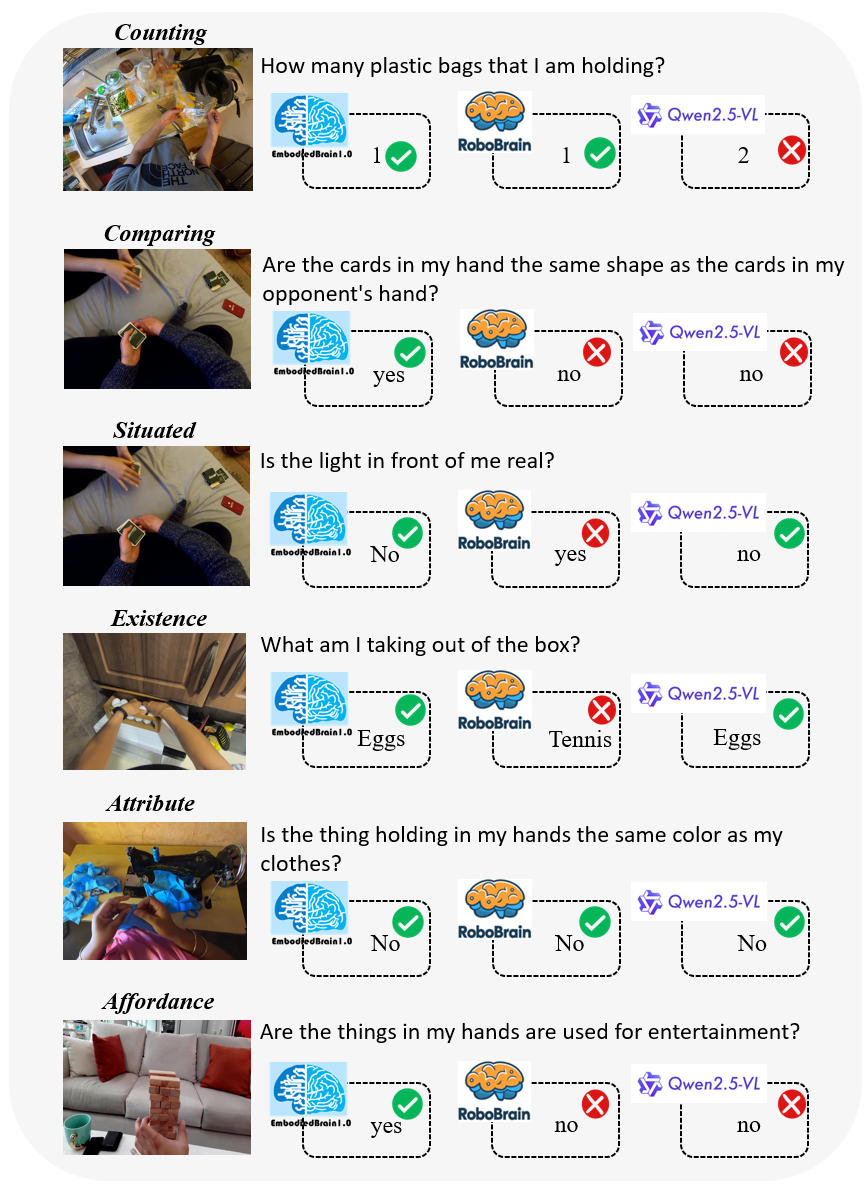}
        \caption{Egothink Example 1}
        \label{fig:judge1}
    \end{subfigure}
    \hfill
    \begin{subfigure}{0.45\linewidth}
        \centering
        \includegraphics[width=\linewidth]{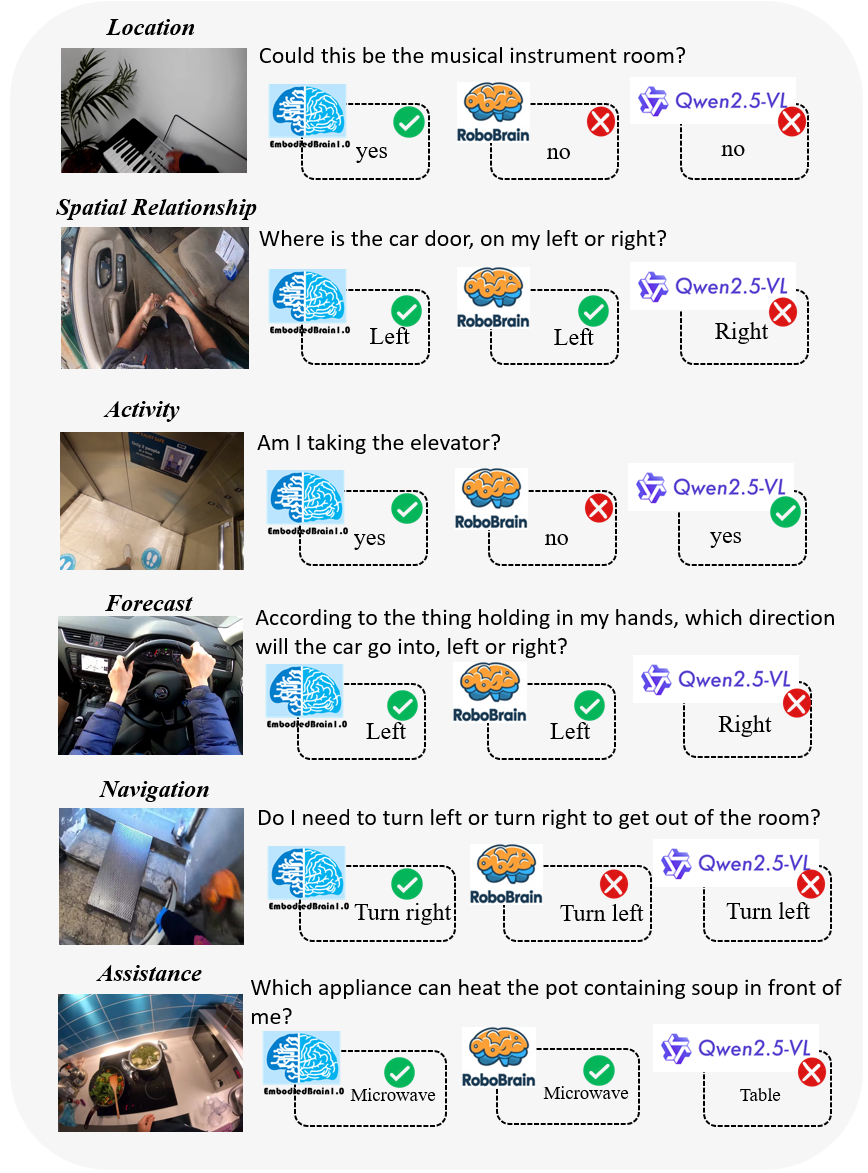}
        \caption{Egothink Example 2}
        \label{fig:judge2}
    \end{subfigure}

    \begin{subfigure}{0.45\linewidth}
        \centering
        \includegraphics[width=\linewidth]{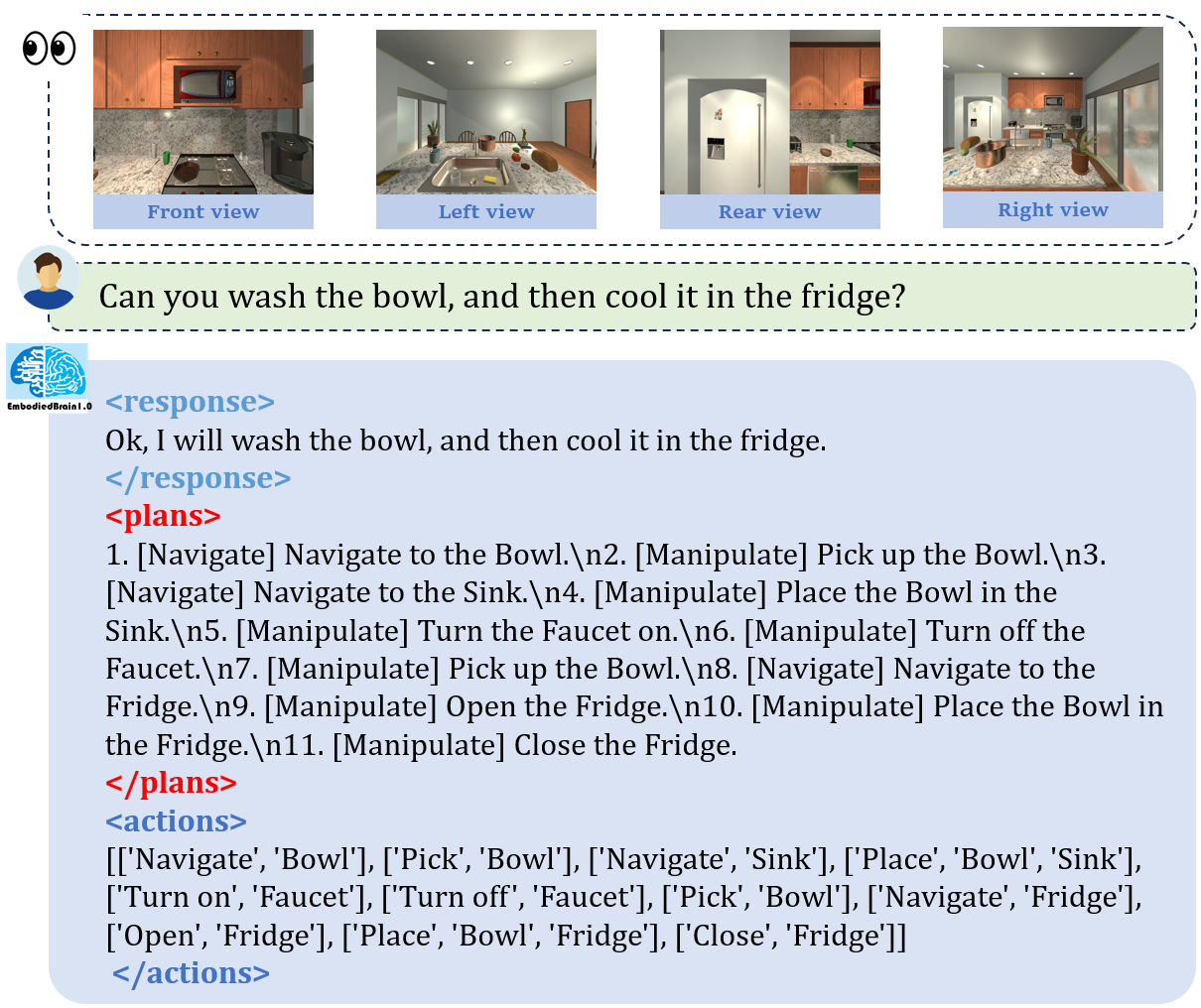}
        \caption{Internal Planing Example 1}
        \label{fig:judge1}
    \end{subfigure}
    \hfill
    \begin{subfigure}{0.45\linewidth}
        \centering
        \includegraphics[width=\linewidth]{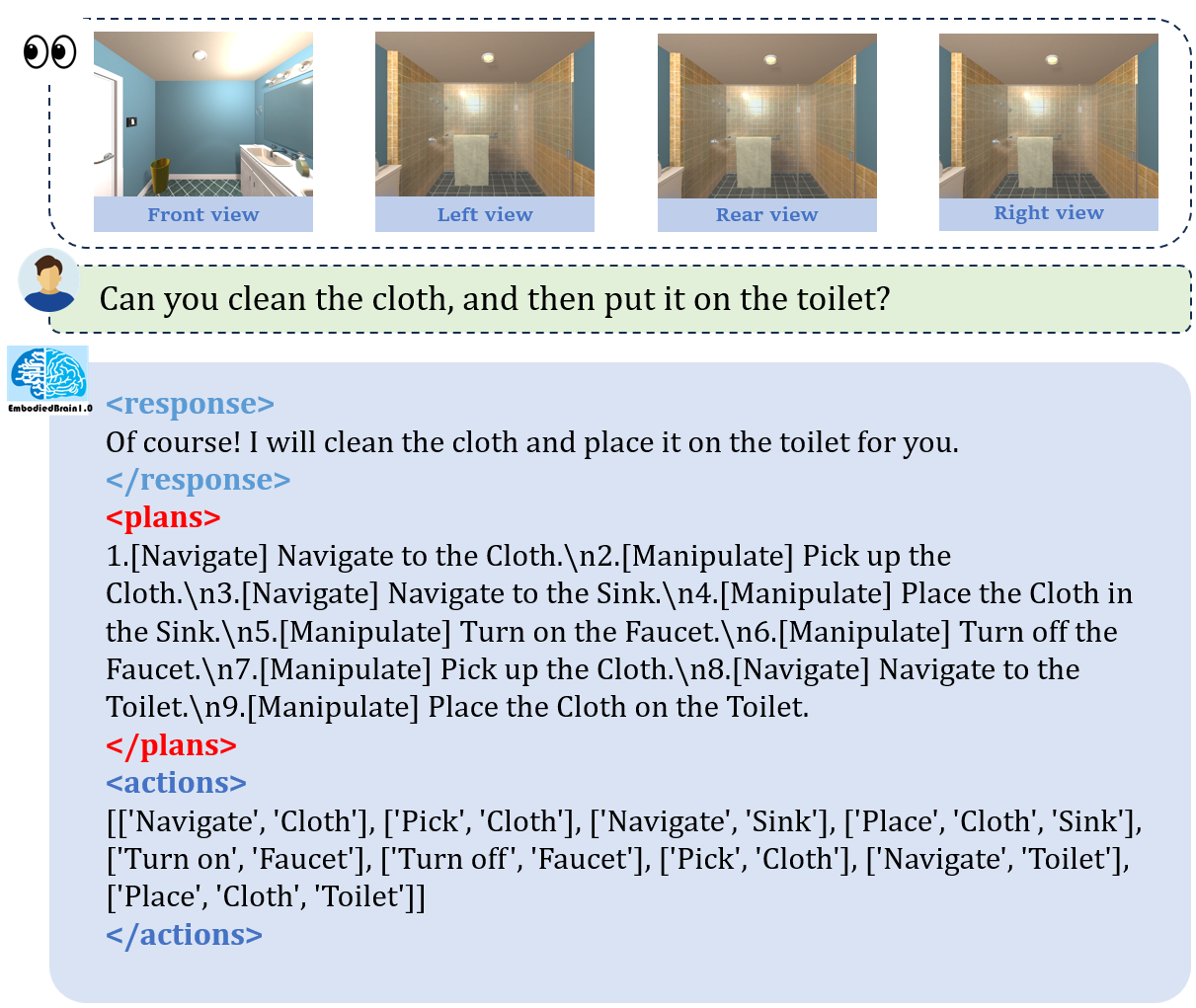}
        \caption{Internal Planing Example 2}
        \label{fig:judge2}
    \end{subfigure}
    
    \caption{\textbf{Examples from EmbodiedBrain: Egoplan and Internal Planing tasks.}}
    \label{fig:all_examples}
\end{figure*}

As shown in Figure~\ref{fig:all_examples}, representative examples from the EmbodiedBrain evaluation suite are presented to illustrate capabilities across different embodied reasoning tasks. The first two subfigures (a–b) are drawn from the Egoplan benchmark, which evaluates an agent’s ability to select the correct next action in a long-horizon task based on a sequence of egocentric video frames. In subfigure (a), given a video depicting progress toward baking pastry and setting a stopwatch, EmbodiedBrain selects the correct option (A) “attach stopwatch,” whereas RoboBrain and Qwen2.5-VL choose incorrect actions (B and C, respectively). In subfigure (b), when asked to loosen a screw from a 3D printer, EmbodiedBrain correctly chooses (D) “loosen screw from 3D printer,” while both baselines select (A) “pick screwdriver,” reflecting a stronger grasp of task state and progression in EmbodiedBrain.

Subfigures (c–d), corresponding to the EgoThink benchmark. The EgoThink benchmark contains 700 question-answer pairs that evaluate six core capabilities: Object, Reasoning, Activity, Forecasting, Localization, and Planning. These are further subdivided into twelve fine-grained dimensions. Subfigure (c) and (d) present sample responses on this benchmark, comparing the outputs of EmbodiedBrain with those of RoboBrain and Qwen2.5-VL. Taking the Navigation item under the Planning task as an example, EmbodiedBrain produced the correct action “Turn Right,” whereas RoboBrain and Qwen2.5-VL selected the incorrect action “Turn Left,” demonstrating the superior task-planning and spatial understanding of EmbodiedBrain.

The final two subfigures (e–f) demonstrate performance on complex, multi-step instruction-following tasks in the internal planing benchmark. Each example provides four egocentric views (front, back, left, right) as input. In subfigure (e), in response to the instruction “Can you wash the bowl, and then cool it in the fridge?”, EmbodiedBrain generates a coherent natural-language response, a detailed step-by-step plan, and a structured sequence of executable actions that correctly navigate, manipulate objects, and interact with appliances. Similarly, in subfigure (f), given the request “Can you clean the cloth, and then put it on the toilet?”, EmbodiedBrain produces a semantically appropriate reply and a precise action plan that successfully completes the task. These examples highlight effective integration of perception, planning, and grounded action generation in realistic simulated environments.

\textbf{\seedblue{End-to-end Sim and VLM-PlanSim-99 benchmarks}}\label{ssec:case_study}

\begin{figure*}[ht]
    \centering
    \includegraphics[width=\linewidth]{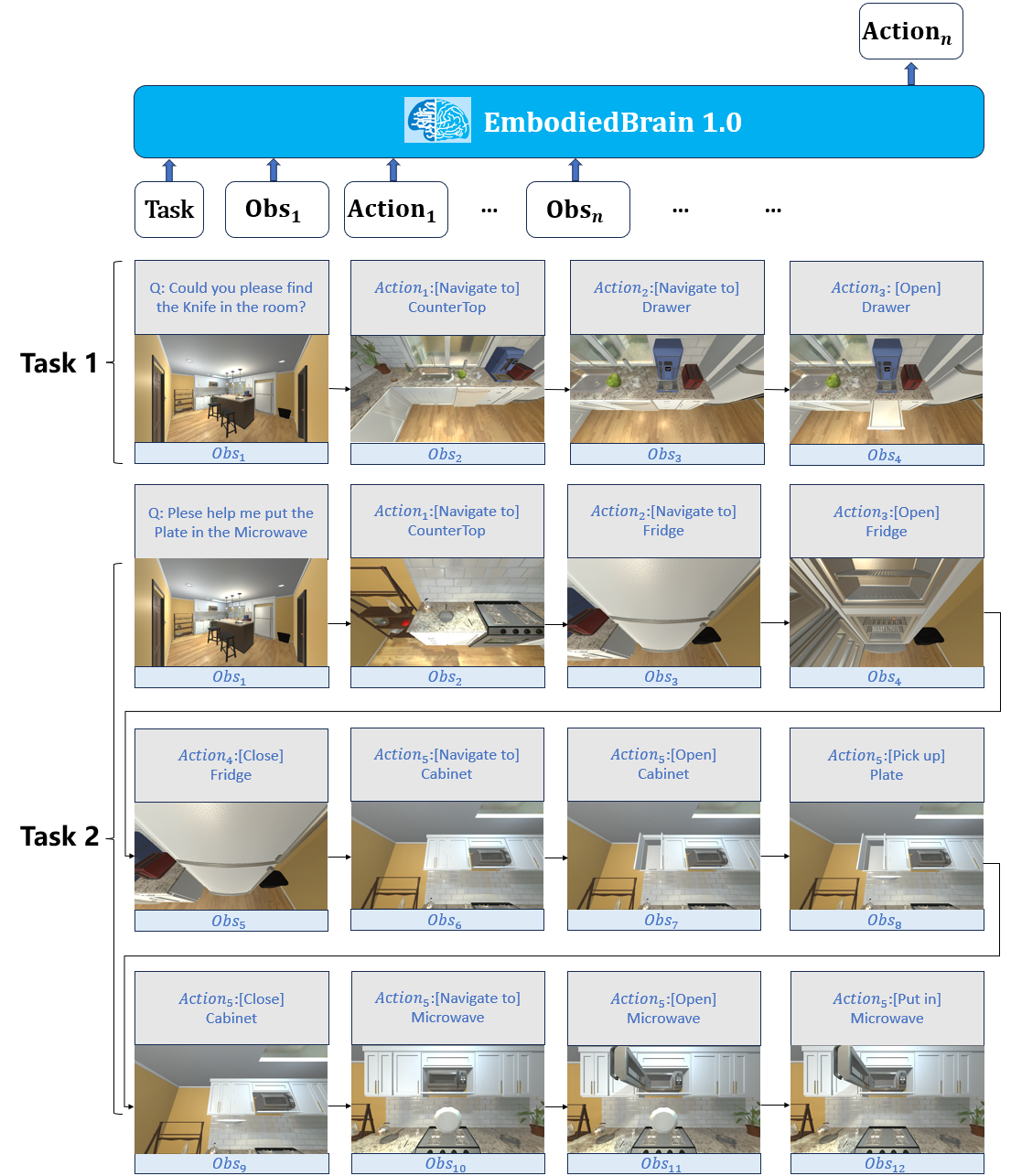}
    \caption{\textbf{End-to-end Sim Examples of EmbodiedBrain.}}
    \label{fig:embodied_reasoner_example}
    \vspace{-1.5em}
\end{figure*}

\begin{figure}[th]
    \centering
    \includegraphics[width=0.95\textwidth]{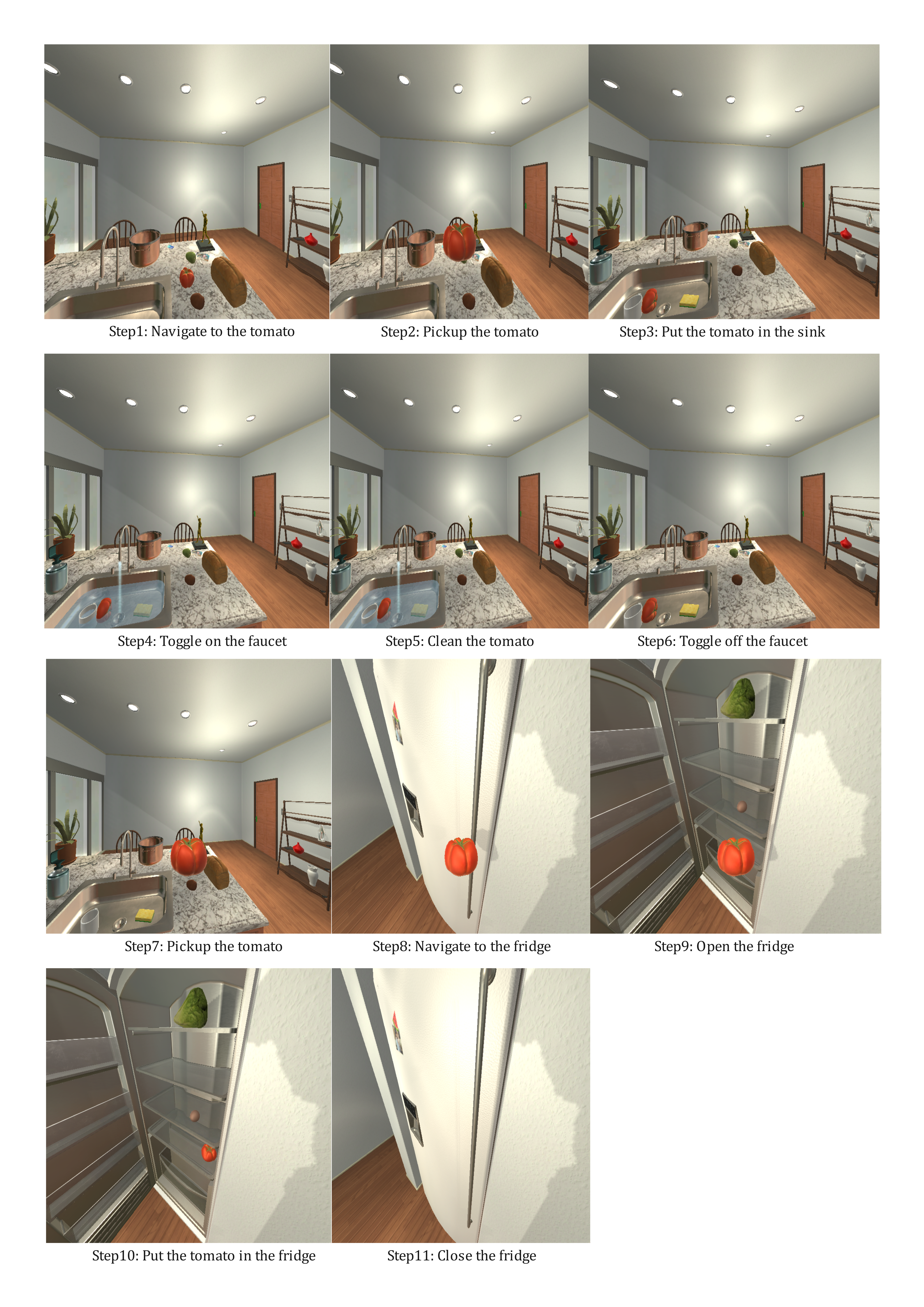}
    \vspace{-20pt}
    \caption{A qualitative example of a successful 11-step plan generated by EmbodiedBrain for the task "Wash the apple in the sink and then put it in the refrigerator." The model correctly identifies and executes two sequential sub-goals: (1) Steps 1-6: acquiring the object, placing it in the sink, and washing it; (2) Steps 7-11: picking up the clean object and storing it in the refrigerator.}
    \label{fig:qualitative_example}
\end{figure}

Figure~\ref{fig:embodied_reasoner_example} illustrates two representative task executions from the EmbodiedReasoner end-to-end simulation platform~\cite{zhang2025embodied}. Task 1 constitutes a short-horizon task, whereas Task 2 represents a long-horizon scenario. In Task 1, EmbodiedBrain rapidly infers that the target object (a knife) is likely located within a drawer and successfully accomplishes the goal in only three action steps. In Task 2—“put the plate in the microwave”—the agent initially hypothesizes that the plate resides in the refrigerator. After opening the refrigerator and failing to locate the plate, EmbodiedBrain dynamically updates its belief, proceeds to inspect a cabinet, successfully retrieves the plate, opens the microwave, and places the plate inside, thereby completing the task in 11 steps. This demonstrates the model's capacity for adaptive reasoning and sequential decision-making in complex, partially observable environments.

In embodied intelligence, a critical measure of an agent's capability is its ability to decompose complex, long-horizon instructions into a coherent sequence of executable actions. This requires not only understanding the explicit goals but also inferring implicit dependencies and state changes. To provide a qualitative illustration of our \textbf{EmbodiedBrain}'s advanced planning capabilities, we present a detailed case study for the high-level command: \textit{“Wash the apple in the sink and then put it in the refrigerator.”}

Figure~\ref{fig:qualitative_example} visualizes the complete 11-step execution sequence generated by our model and validated in our AI2-THOR framework. The model correctly segments this complex goal into two primary sub-tasks: (1) washing the object, and (2) storing it. As the sequence demonstrates, the agent exhibits robust, fine-grained reasoning. It begins by navigating to and acquiring the target object (\textit{Steps 1-2}). It then correctly identifies the sink as the necessary location for the "wash" action, places the apple in it, and interacts with the environment (toggling the faucet) to achieve the 'cleaned' state (\textit{Steps 3-6}). Once the first sub-task is complete, the agent re-acquires the now-clean apple (\textit{Step 7}) and proceeds to the second goal. This involves navigating to the refrigerator, opening it, placing the apple inside, and closing the door to secure the final state (\textit{Steps 8-11}). The flawless, end-to-end execution of this plan highlights our model's strong performance in sequential reasoning, state tracking, and grounding abstract linguistic concepts (like "wash") into physically-sound interactions.

\clearpage

\subsection{Prompts Details}

\subsubsection{Prompts for Spatial Rejection Sampling}

\begin{tcolorbox}
[colback=white,colframe=black,left=1mm,right=1mm,top=1mm,bottom=1mm,breakable]
\ttfamily

You are an evaluation assistant (consistency checker).\par
Your task:\par
1. Receive a question (Question) and two responses (Answer A / Answer B).\par
2. Determine whether the two responses are semantically consistent or inconsistent.\par
3. Output strictly in JSON format:\par
\ \ \{"consistent": true\_or\_false, "reason": "..."\}\par
\ \ - consistent=true means the responses are semantically equivalent or non-conflicting.\par
\ \ - consistent=false means the responses contradict or are inconsistent in meaning.\par
4. Return only the JSON object with no additional text or explanation.

\end{tcolorbox}

\subsubsection{Prompts for Video Understanding Multiple-Choice Reasoning Generation}
\label{vu_reasoning_prompt}
\begin{tcolorbox}[colback=white,colframe=black,left=1mm,right=1mm,top=1mm,bottom=1mm]
\ttfamily
You are an expert video analysis assistant specializing in action prediction and temporal reasoning.

Your task:

1. Analyze the video progress and final frame observation to determine the optimal next action for accomplishing: \{task\}

2. Evaluate the provided answer choices (A1-A4) based on multi-modal reasoning: \{choices\}

3. Output strictly in JSON format:

\{ \\
\ "choose\_answer": "A1",\\
\ "reasoning\_chain": \{\\
\ \ "video\_context\_analysis": "...",\\
\ \ "final\_frame\_observation": "...", \\
\ \ "temporal\_progression\_assessment": "..." \}, \\
\ "comparative\_evaluation": \{\\
\ \ "selected\_answer\_strengths": "...",\\
\ \ "rejected\_answers\_weaknesses": ["..."] \}\\
\}

Return only the JSON object with no additional text.
\end{tcolorbox}

\subsubsection{Prompts for Planning data Generation}
\begin{tcolorbox}
[colback=white,colframe=black,left=1mm,right=1mm,top=1mm,bottom=1mm,breakable]
\ttfamily

You are a professional data format conversion expert tasked with transforming a planning data description into the format I specify. Please strictly follow the requirements and examples below:\par
\medskip
\textbf{Original Data Structure}\par
1. A textual task planning (plan) sequence\par
\medskip
\textbf{Conversion Task}\par
You must perform the following transformations:\par
\medskip
1. \textbf{Plan Enhancement Requirements}:\par
\ \ - Add an atomic action tag to each plan step (format: [Action])\par
\ \ - Select the most appropriate action from the following atomic action list: \{", ".join(prime\_actions)\}\par
\ \ - Example: "1. Navigate to the First Cargo Section" becomes "1.[Navigate] Navigate to the First Cargo Section"\par
\medskip
2. \textbf{Actions Generation Rules}:\par
\ \ - Format: A Python-style 2D list: [['Action', 'Object1'], ['Action', 'Object1', 'Object2'], ...]\par
\ \ - Action selection: You must choose actions only from the provided atomic action list. If no exact match exists, select the closest available action.\par
\ \ - Object conventions:\par
\ \ \ \ 1. Use concise English noun phrases (e.g., "Cargo Straps")\par
\ \ \ \ 2. Single-object actions: ['Action', 'Object'] (e.g., ['Find', 'Cargo Straps'])\par
\ \ \ \ 3. Two-object actions: ['Action', 'Object', 'Target'] (e.g., ['Put', 'Bread', 'Basket'])\par
\medskip
\textbf{Strictly Prohibited Actions}\par
1. Altering the core semantics of the original plans\par
2. Adding explanatory text or comments\par
3. Using actions not included in the atomic action list\par
4. Changing the core reference of any object\par
\medskip
---\par
\medskip
\textbf{Correct Conversion Example:}\par
<plans>1.[Navigate] Navigate to the First Cargo Section\textbackslash n2.[Find] Visually Inspect the Cargo Straps for Tightness\textbackslash n3.[Adjust] Adjust any Loose Straps\textbackslash n4.[Navigate] Move to the Next Cargo Section</plans><actions>[['Navigate', 'First Cargo Section'], ['Find', 'Cargo Straps'], ['Adjust', 'Loose Straps'], ['Navigate', 'Next Cargo Section']]</actions>\par
\medskip
\textbf{Incorrect Conversion Example:}\par
<plans>1.[Navigate] Navigate to the First Cargo Section\textbackslash n2.[Find] Visually Inspect the Cargo Straps for Tightness\textbackslash n3.[Adjust] Adjust any Loose Straps\textbackslash n4.[Navigate] Move to the Next Cargo Section\textbackslash n5.[Repeat] Repeat the Inspection and Adjustment Process\textbackslash n6.[Continue] Continue Until All Sections are Checked</plans><actions>[['Navigate', 'First Cargo Section'], ['Find', 'Cargo Straps'], ['Adjust', 'Loose Straps'], ['Navigate', 'Next Cargo Section'], ['Repeat', 'Inspection and Adjustment Process'], ['Continue', 'All Sections']]\par
\medskip
\textbf{Reasons for error}:\par
1. Missing closing </actions> tag\par
2. The atomic action list does not contain the action 'Continue'\par
\medskip
---\par
\medskip
Now, please process the following data and output \textbf{only} the final converted result.

\end{tcolorbox}

\subsubsection{Prompts for Generative Reward Model in Step-GRPO} \label{GRM_prompt}
\begin{tcolorbox}
[colback=white,colframe=black,left=1mm,right=1mm,top=1mm,bottom=1mm,breakable]
\ttfamily

You are an \textbf{expert reviewer in the field of robotic task response and planning}. Your task is to provide a \textbf{rigorous, objective, and fair evaluation} of the model-generated response, task plan, and action sequence. Scores must range from 0.00 to 1.00, with exactly two decimal places. Please provide a brief reasoning before assigning the score.\par
\medskip
---\par
\medskip
\textbf{Scoring Guidelines}\par
Evaluate the model's output against the reference answer based on, but not limited to, the following four core dimensions:\par
\medskip
---\par
\medskip
\textbf{1. <response> Response Quality}\par
- \textbf{Politeness}: Does it naturally acknowledge the user request with a friendly tone?\par
- \textbf{Accuracy}: Does it correctly reflect the task intent?\par
- \textbf{Conciseness}: Is it clear and free of redundancy?\par
- \textbf{Naturalness}: Does the language sound fluent and idiomatic?\par
\medskip
---\par
\medskip
\textbf{2. <plans> Planning Quality}\par
- \textbf{Completeness}: Does it cover \textbf{all necessary steps} from start to task completion?\par
- \textbf{Logical Order}: Is the step sequence reasonable? (e.g., no "manipulate before navigate" errors)\par
- \textbf{Executability}: Is each step grounded in visible scene content, without assuming unseen objects?\par
- \textbf{Formatting Compliance}:\par
\ \ - Are only the two uppercase tags \texttt{[Navigate]} and \texttt{[Manipulate]} used strictly?\par
- \textbf{Redundancy}: Are there any meaningless or repetitive steps?\par
\medskip
---\par
\medskip
\textbf{3. <actions> Action Sequence Quality}\par
- \textbf{Verb Accuracy}: Are all verbs in <actions> \textbf{exclusively} from the predefined atomic action set? Any verb outside this set results in penalty.\par
- \textbf{Structural Consistency}:\par
\ \ - Is the semantic meaning aligned with <plans>?\par
- \textbf{Alignment with <plans>}: Does the <actions> sequence semantically match the <plans> steps, using consistent object and location names?\par
- \textbf{Argument Completeness}: Are required objects and locations included?\par
- \textbf{Order Consistency}: Does the action sequence correspond one-to-one with the <plans> steps?\par
\medskip
---\par
\medskip
\textbf{4. Format Compliance}\par
- \textbf{Overall Structure}: Does it strictly follow the XML format: <response>...</response>
<plans>...</plans><actions>...</actions>, with all tags properly closed and no extra/missing parts?\par
- \textbf{Nesting and Order}: Are the tags correctly nested and sequenced?\par
\medskip
---\par
\medskip
\textbf{Critical Error Criteria (trigger low scores if any apply)}\par
- Use of undefined action verbs.\par
- Reference to objects or locations not present in the reference answer.\par
- Missing critical steps (e.g., failing to grasp an item or deliver to user).\par
- Severe action ordering errors (e.g., placing before grasping).\par
- Invalid output format (e.g., missing XML tags, malformed JSON).\par
\medskip
---\par
\medskip
\textbf{Scoring Rubric (for reference)}\par
- \textbf{1.00}: Nearly identical to reference; natural language, flawless logic.\par
- \textbf{0.90–0.99}: Semantically equivalent; minor phrasing differences, no logical errors.\par
- \textbf{0.75–0.89}: Reasonable structure; slight omissions or redundancies, but still executable.\par
- \textbf{0.50–0.74}: Minor logical or formatting issues; partially executable.\par
- \textbf{0.25–0.49}: Clear errors (e.g., wrong order, missing actions).\par
- \textbf{0.00–0.24}: Severe errors (e.g., illegal actions, nonexistent objects, format collapse).\par
\medskip
---\par
\medskip
\textbf{Original Question}\par
\{question\}\par
\medskip
---\par
\medskip
\textbf{Reference Answer}\par
\{sol\}\par
\medskip
---\par
\medskip
\textbf{Atomic Action Set}\par
\{ATOMIC\_ACTION\_SET\}\par
\medskip
---\par
\medskip
\textbf{Model-Generated Output}\par
\{completion\}\par
\medskip
---\par
\medskip
\textbf{Please place your final score inside <score></score> tags. Include an extremely concise reasoning inside <think></think> tags. \textbf{Do not provide lengthy or detailed analysis.}}\par
\medskip
Example response: <think>Brief reasoning</think><score>0.75</score>

\end{tcolorbox}

\clearpage

\bibliographystyle{plainnat}
\bibliography{main}


\end{document}